\documentclass[lettersize,journal]{IEEEtran}

\IEEEoverridecommandlockouts                              % This command is only needed if 
\usepackage{graphicx}
\usepackage{amsmath}
\usepackage{amssymb}
\usepackage[english]{babel}
\usepackage[utf8]{inputenc}
\usepackage{algorithm}
\usepackage[noend]{algpseudocode}
\usepackage{cleveref}
\usepackage{times}
\usepackage{mathtools}
\usepackage{booktabs}
\usepackage{bm}
\usepackage[dvipsnames]{xcolor}
\usepackage{tikz}
\usepackage{caption}
\usepackage{gensymb}
\usepackage{subcaption}
\usetikzlibrary{arrows,shapes,automata,backgrounds,petri}
\usepackage[EULERGREEK]{sansmath}

\usepackage{pgfplots}
\pgfplotsset{compat=newest}
\usetikzlibrary{plotmarks}
\usetikzlibrary{arrows.meta}
\usepgfplotslibrary{patchplots}
\usepackage{grffile}

\newcommand{\bh}{\bm{h}}

\newcommand{\bx}{\bm{x}}
\newcommand{\by}{\bm{y}}

\newcommand{\cN}{\mathcal{N}}

\newcommand{\cT}{{\mathcal{T}}}

\newcommand{\cX}{\mathcal{X}}
\newcommand{\cY}{\mathcal{Y}}

\newcommand{\EE}{\mathbb{E}}

\newcommand{\RR}{\mathbb{R}}

\newcommand{\blambda}{\bm{\lambda}}

\newcommand{\maximize}{\mathop{\mathrm{max}}}
\newcommand{\minimize}{\mathop{\mathrm{min}}}
\newcommand{\maximizewrt}[1]{\mathop{\underset{#1}{\maximize}~}}
\newcommand{\minimizewrt}[1]{\mathop{\underset{#1}{\minimize}~}}

\newcommand{\norm}[1]{\left\| #1\right\|}

\newcommand{\defeq}{\vcentcolon=}

\newcommand{\Expect}[2]{\EE_{#1}\left[#2\right]}

\title{\LARGE \bf
Learn from Human Teams: a Probabilistic Solution to Real-Time Collaborative Robot Handling with Dynamic Gesture Commands
}

\author{Rui Chen$^{1*}$, Alvin Shek$^{1*}$, Changliu Liu$^{1}$% <-this % stops a space
\thanks{*Equal contribution.}% <-this % stops a space
\thanks{$^{1}$Carnegie Mellon University, Pittsburgh, PA. Contact: {\tt\small \{ruic, ashek, cliu6\}@andrew.cmu.edu}}%
}

\begin{document}

\maketitle
\thispagestyle{empty}
\pagestyle{empty}

%%%%%%%%%%%%%%%%%%%%%%%%%%%%%%%%%%%%%%%%%%%%%%%%%%%%%%%%%%%%%%%%%%%%%%%%%%%%%%%%

% Use the math macros in \textbf{math\_macros.tex} if possible, for example $\bR, \Rb, \RR, \cR, \Expect{q(z)}{p(x,y,z)}, \st, \partialwrt{x}$, and others.

\begin{abstract}

We study real-time collaborative robot (cobot) handling, where the cobot maneuvers a workpiece under human commands. This is useful when it is risky for humans to directly handle the workpiece. However, it is hard to make the cobot both easy to command and flexible in possible operations. In this work, we propose a \textit{Real-Time Collaborative Robot Handling} (RTCoHand) framework that allows the control of cobot via user-customized dynamic gestures. This is hard due to variations among users, human motion uncertainties, and noisy human input. We model the task as a probabilistic generative process, referred to as \textit{Conditional Collaborative Handling Process} (CCHP), and learn from human-human collaboration. We thoroughly evaluate the adaptability and robustness of CCHP and apply our approach to a real-time cobot handling task with Kinova Gen3 robot arm. We achieve seamless human-robot collaboration with both experienced and new users. Compared to classical controllers, RTCoHand allows significantly more complex maneuvers and lower user cognitive burden. It also eliminates the need for trial-and-error, rendering it advantageous in safety-critical tasks.

\end{abstract}

%%%%%%%%%%%%%%%%%%%%%%%%%%%%%%%%%%%%%%%%%%%%%%%%%%%%%%%%%%%%%%%%%%%%%%%%%%%%%%%%

\section{INTRODUCTION}
\subsection{Material Handling with Cobots}
With the advancement of robotic technologies, robots are getting out of cages and directly working with humans. One immediate application of collaborative robot (\textit{cobot}) is material handling where the robot moves or presents a workpiece under human commands in real time. It is particularly useful when the material to handle is heavy, as in automotive assembly, or when the material needs to stay untouched by human, as in the case of food production \cite{hagele_industrial_2016}, or when there are safety risks, as in the case of dangerous liquid \cite{ende_human_centered_2011}. Using cobots for handling also improves product lifecyles and customization \cite{villani_survey_2018}. Factory workers identify material handling as an aspect where cobots could help~\cite{attitude_2017}.

% This paper targets robot handling. Refer to HRC survey for motivation (although limited intelligence, still adv because of natural work flow; human cannot handle, dangerous). Also refer to Julie Shah paper about workers' altitude towards robot helper (robots just do what humans cannot, not intelligence part).

The choice of human interface is key to making cobots easy to use. There are two major types of interfaces for interacting with robots. The first one is \textit{classical devices} (e.g., keyboards and joysticks) which requires either tedious integration efforts or programming expertise and can hardly be used in real-time. The second one is \textit{natural user interfaces} (NUIs) (e.g., gestures and voices) which requires no technical skills and can naturally be invoked in real-time tasks. For cobot handling, we choose NUIs for usability. In human-robot collaboration (HRC) tasks where human workers work with robots to achieve common goals, static gesture is one of the most studied NUIs \cite{ende_human_centered_2011,shukla2016,mazhar_real_time_2019,gesture_hri_review,simao_online_2019,marasovic_motion_based_2015}. However, in real-time cobot handling where the desired material movements change continuously, static gesture would show limited flexibility due to its discrete nature. For example, when inspecting a workpiece, the user might keep rotating the workpiece towards any directions for desired view angles. This incurs an infinite number of possible operations and cannot be represented by finite static gestures. Hence, we propose to use \textbf{dynamic gesture}, since it enables users to directly mimic or depict a mental image of desired material movements, leading to both flexibility and fluidity of interaction. With that in hand, an immediate question is: how to make cobots understand the dynamic gestures from different human users in various tasks, and react with handling operations that meet humans' expectation? This paper aims to address this problem.

% To enable robot handling, interface is important. Several existing methods refering to section 4 of HRC review. Summarize downsides (have to contact robot, non-intuitive interface, etc.). Propose hand control.

\begin{figure}
    \centering
    \includegraphics[width=\linewidth]{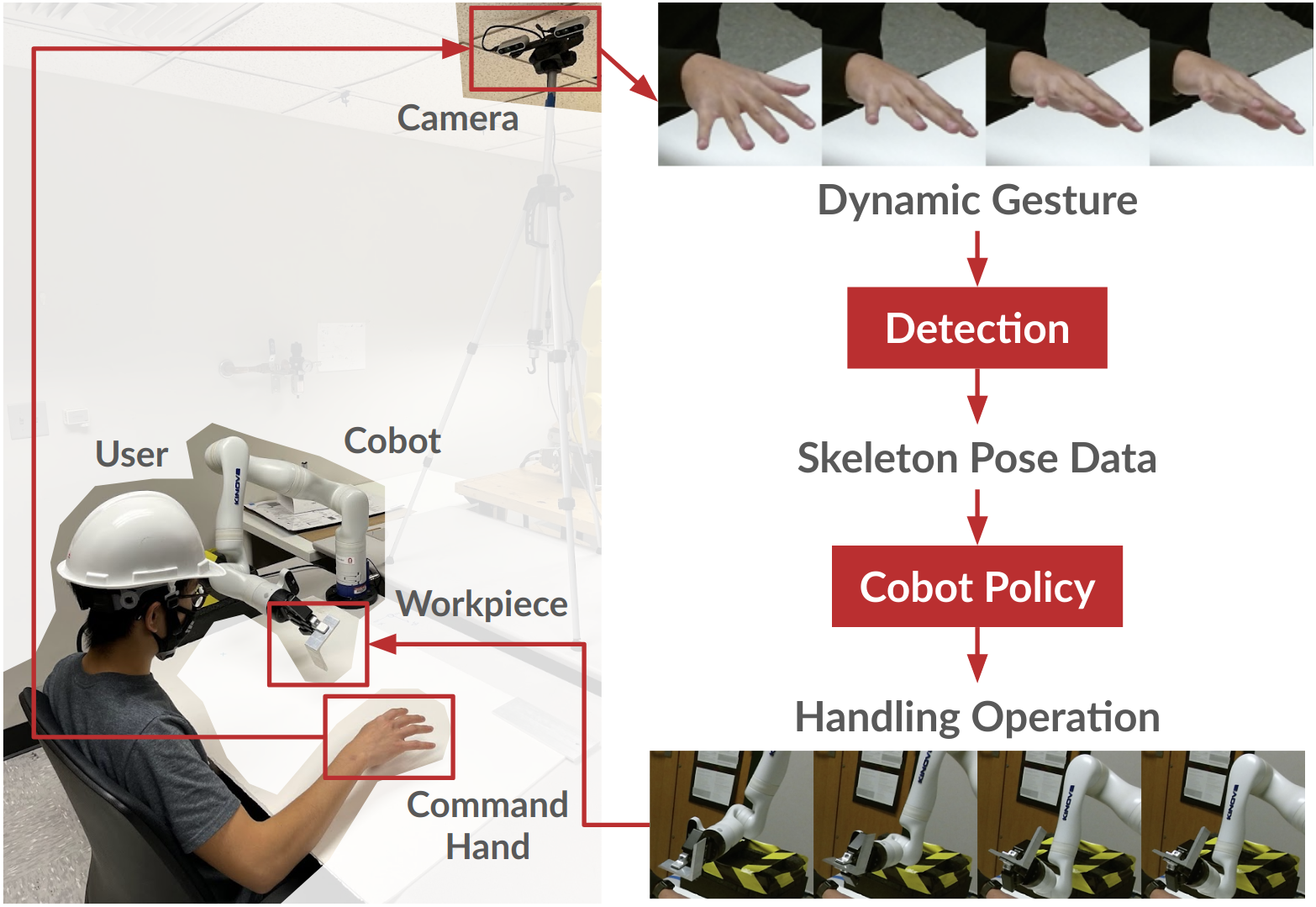}
    \caption{Illustration of a collaborative robot (cobot) handling task. Users control the workpiece handling operations using only dynamic gestures as commands. The dynamic gestures are first captured by multi-view cameras and then processed using skeleton detection toolbox. Then, a cobot policy maps user commands to handling operations, which are finally executed by the cobot in real time.}
    \label{fig:task_diagram}
\end{figure}

\subsection{A Real-Time Collaborative Robot Handling Framework}
One challenge in generic HRC tasks comes from the trade-off between accessibility and flexibility. An HRC task is accessible if it demands little technical skill, cognitive efforts, and physical efforts for the user to interact with the cobot. The task is flexible if the user can perform a wide range of operations as desired. For example, both tele-operation via teaching pendant and remote simulation \cite{olp_2013} enable flexible tasks and accommodate dangerous materials, but are not highly accessible because they require non-intuitive programming skills \cite{villani_survey_2018} and can quickly be tedious as tasks become more complex. On the other hand, walk-through programming (also known as lead-through teaching) is highly accessible because it allows users to physically move the robot end-effector and requires no programming knowledge. However, it is not suitable for tasks that forbid human-robot physical contact and as a result, not flexible. In material handling, for example, the robot end-effector might not be reachable by users when a bulky workpiece is hung on it, or when the handled material might lead to health risks.

To achieve both high accessibility and flexibility, we propose a novel \textit{\textbf{R}eal-\textbf{T}ime \textbf{Co}llaborative Robot \textbf{Hand}ling} (RTCoHand) framework. This framework enables users to generate complex and flexible workpiece trajectories via only \textit{dynamic gestures} without any physical contact with the cobot or workpiece. To maintain high accessibility to different users, we introduce \textit{policy customization} which allows users to customize their own dynamic gestures as commands for arbitrary desired handling operations.
% While this feature addresses the issue of physical capabilities, the gesture design process can be demanding due to the number of possible handling operations. As such, we include two other features in this framework, \textit{dominant motion decomposition} and \textit{reference initial gesture}, to reduce users' cognitive burden while preserving the complexity of workpiece trajectories. 
In later sections (e.g., section \ref{sec:rt_cobot_handling_task}), we will verify that RTCoHand provides a practical setting for cobot handling tasks with both high accessibility and flexibility.

% In our human study, all \ruic{\#} human subjects agree that (a) there is no obvious alternative to the provided dynamic gesture reference for better ergonomics, (b) they can easily manage to combine dominant motions for versatile robot handling operations without prior technical knowledge, and (c) it is physically hard or even infeasible to command some general handling motions using a single dynamic gesture. 

% Challenge: dilemma of accessibility and flexibility. Solution: novel robot handling task specification that both allow maximum possible operations and require no robot programming knolwedge and minimum cognitive burden on workers.
\subsection{Enabling Adaptability Using Conditional Cobot Handling}
Another challenge in designing robot policies in HRC tasks is how to adapt to different tasks \cite{villani_survey_2018}, different users styles \cite{attitude_2017}, and uncertainty of natural human input. For example, users might have different joint flexibilities which would render certain gestures natural for some users but hard for others. Allowing users to develop their own gestures mitigates this issue but raises challenges for cobot's adaptivity. To enable the cobot to adapt to each user, we make the cobot aware of the past working experience with that same user and use it as a reference for interpreting new commands. Besides, natural human inputs such as dynamic gestures can be noisy due to human uncertainty and limited quality of the sensing systems. To capture such uncertainty, we learn and maintain a distribution of cobot policies. Finally, for safe collaboration without abrupt movements, the robot needs to robustly generate smooth actions even when the human input is noisy. To achieve that, we explicitly encourage the continuity of cobot actions in policy design.

Integrating the ideas mentioned above, we frame the problem of cobot handling as learning a \textit{distribution of functions}, where each realization is a policy variant that maps real-time human dynamic gestures to cobot end-effector motions. The adaptation to different users is solved via conditional dependency on user-specific demonstrations. The uncertainty of human input is naturally captured by various policy variants. The smoothness of cobot handling is achieved by a set of conditional probabilities that encourages dependency between consecutive cobot actions. To learn the distribution, we propose {\em Conditional Collaborative Handling Process} (CCHP) inspired by a recent line of research on neural processes (NP) \cite{cnp, np, anp}. To ease the learning of long handling operations, we deploy teacher forcing technique \cite{bengio2015teacherforce} during training. As a result, the model initially learns short-term policies based on labeled trajectories and gradually disregards labels to learn long-term policies. We train our cobot policy model and show that the cobot can adapt to user-specific characteristics such as general hand gestures (e.g., whether the user prefer to keep fingers extended or retracted), scale of hand motions, and rotation styles (e.g., whether users rotate hands in-place or while moving the hands). The cobot can also generate reasonable actions in a few-shot fashion, i.e., when the desired handling operation has not been directly labeled for the user, but is inferred with insights drawn from relevant user demonstrations. We implement the RTCoHand framework with a Kinova Gen3 robot arm and verify our approach on a real-time cobot handling task: hot metal inspection. Our user study shows that RTCoHand framework significantly out-performs classical controller-based approach in terms of more complex maneuvers, lower user cognitive burden, and no need for trial-and-error.

\subsection{Overview of Contributions and Conclusions}

In short, we summarize our contribution as follows. First, we propose \textbf{RTCoHand}, a novel real-time collaborative robot handling framework where users can achieve complex workpiece maneuvers using self-designed dynamic gestures. Then, to acquire the cobot policy, we propose \textbf{Conditional Collaborative Handling Process (CCHP)}, a probabilistic generative process that models cobot handling tasks. CCHP enables across-task and across-user adaption, accommodates human uncertainty, and generate robust cobot actions against noisy input. Finally, we demonstrate the application of RTCoHand framework on a \textbf{real-time handling task}, collaborative hot metal inspection, with a physical robot. With RTCoHand, we achieve seamless human robot collaboration with complex workpiece maneuvers and low user cognitive burden. Users are able to confidently control the cobot actions without trial-and-error, meaning that RTCoHand is advantageous in safety-critical scenarios where mistakes are to be avoided.

% \ruic{Intro to rest of paper is omitted for now due to paper length.}

% too less or too much guidance could both burden users with designing moves from scratch or memorizing the designed moves

% Challenge: (1) refer HRC paper, consider different skill levels and worker behaviors. (2) refer to Julie Shah human-robot interface, human workers are heterog, work at diff speed, not all team members can do same thing. Congnitive demands is still heavy. Summarize challenges: variations within each worker and among diff workers. (3) explainability and human trust. Solution: main model (solves both adaptability and explainability).Summarize techniques, most important base model and dev (dependency and teacher forcing)

\section{Related Work}

% \subsection{HRC interfaces}

% classical approaches, novels such as walk-though, offline, multimodal and why they are not suitable for our application.

\subsection{Gesture-Based Human Robot Interaction}

In human-robot interaction, hand gesture has been widely considered as an intuitive tool for human users to communicate with robots. In literature, there are two major forms of gesture-based commands: \textit{static gestures} and \textit{dynamic gestures}. Regarding static gestures, the most common approach is to classify gesture poses using a finite set of symbolic labels. These labels are further mapped to robot actions in rule-based human-robot interactions, such as handling of dangerous liquid \cite{ende_human_centered_2011} and teach and replay \cite{mazhar_real_time_2019}. Note that static gesture is not suitable for cobot handling tasks, since there are infinitely many possible handling operations which cannot be matched by a finite set of labels.

Regarding dynamic gestures, \cite{marasovic_motion_based_2015} treats the whole hand as a single point and interprets its trajectory using some simple geometries such as circle and alphabetical letters. This approach still operates on a discrete set of robot actions, and does not introduce more flexibility on possible operations. We argue that, to fully leverage the expressiveness of dynamic gestures and achieve seamless human-robot interaction, the robot should react to human gestures in real time with continuous actions. Such setting introduces significant challenge in cobot policy because of the continuous robot action space, and is rarely seen in literature. There is one work \cite{hand_to_end_effector} mapping hand motions to real-time robot gripper actions based on hand keypoint detection. Such approach is only applicable when hand-robot correspondence is obvious. It also enforces a narrow range of hand motions which may appear hard to users with limited strength and joint flexibilities. Similarly, \cite{fan_object_2016} achieves real-time object pose tracking from hand motions. Their approach requires costly motion capture systems and wearable markers and hence is not flexible. Also, they do not consider different hand motion styles from different users. In this paper, we desire a cobot handler that can be easily commanded by various users in real time without wearing any device. This requires the cobot to adapt to different user control strategies and hand gesture patterns, which is not solved by existing approaches.

% \ruic{contd}
% Omit recognition \cite{gesture_hri_review}.

% Regarding hand control, summarize existing work using gestures. (1) gesture recog + symbolic encoding: limited flexibility considering motion path and velocity. (2) direct motion mapping: limited application (gripping only). None of existing work handles general handling, i.e., rigid body transformation. Descibe main diff of our work: continuous hand motion. Adv: intuitive, self-designed, no specific skills required.

\subsection{Functional Distribution with Efficient Inference}

The goal of cobot handling is to learn a policy distribution that models human uncertainty, while being able to adapt to different users based on user-specific demonstrations when interpreting new commands. On an abstract level, this is equivalent to regressing a functional distribution which predicts function values at unobserved input locations with uncertainty, given some observations. One direct approach is to perform inference on a stochastic process such as Gaussian process (GP). Although GPs carry all mentioned properties, their computation time scales cubically with respect to the number of observations in original formulation, and quadratically with approximation \cite{gp_quad}. This renders GPs infeasible for real-time tasks which require fast online computations. Moreover, GPs require explicitly defined kernel functions, which can hardly be designed manually for high-dimensional tasks like ours.

There is a line of research that models stochastic processes with a class of neural networks, named \textit{neural processes} (NP), to achieve linear computational complexity with respect to observations during test time. This approach is first formally presented as conditional neural processes (CNP) \cite{cnp} which explicitly incorporate training data at test time as observations. Based on CNP, neural processes (NP) \cite{np} introduces a latent variable to capture the global uncertainty in target functions. Attentive neural processes (ANP) \cite{anp} mitigates the under-fitting issues in NPs via an additional attention module on observations. Although NPs, especially ANP, are effective on tasks such as image completion, they cannot be directly apply to cobot handling which is significantly more complex. In image completion, NPs map a set of 2D image coordinates to RGB pixel values, while in cobot handling, we need to map a sequence of gestures (21D hand skeleton model \cite{simon2017hand}) to 6D object poses in Cartesian space. More importantly, our task output (handling operations) should have strong dependencies between adjacent time steps. Although the latent variable in (A)NP introduces correlation between predictions, such correlation is invariant to the input order and does not consider temporal structures.

To define stochastic processes, we indeed need to ensure invariance to input permutations, i.e., exchangeability condition \cite{Oksendal2003}. However, it is reported to be practically beneficial to relax such assumption when the observations contain time sequences \cite{qin_recurrent_2019}. Specifically, recurrent attentive neural processes (RANP) \cite{qin_recurrent_2019} incorporates a recurrent neural network structure to process the observations, and show improved performance on vehicle trajectory predicition over ANP. We point out that in RANP, the exchangeability is only relaxed on observations, while the temporal structure of test-time input and output is not considered. In cobot handling, we also need to relax exchangeability condition at test time to ensure smooth and consistent cobot actions. Finally, there are other extensions to NPs whose problem setting deviates from ours. For example, recurrent neural processes (RNP) \cite{rnp} focuses on the dynamics of latent variable, while in our task, we assume that to be fixed during the same handling session with the same user. Moreover, sequential neural processes (SNP) \cite{snp,robust_snp} studies the dependency in a sequence of stochastic processes. In our setting, the processes for different users in different sessions are independent.

% and sequential neural processes (SNP) \cite{snp,robust_snp}.

% ANP (RANP, RNP, SNP), GP refer to ANP review

% Few shot for seq forecasting. Our work is regarded as knowing partial sequence. \textbf{important related work} Does it consider uncertainty?

% RNP, SNP considers dynamic latent while in our case the latent z is stationary given a context. RNP has multiple patterns in different scales in one sequence. SNP considers transitions between different sequences. Namely, different motion commands from the same user are not correlated (unlike a changing perspective where each frame is a function and different frames are correlated), while different time steps in a single motion command is still correlated.

\section{Real-Time Cobot Handling Framework (RTCoHand)}\label{sec:rtcohand}

In this section, we first model how users work with the cobot to handle an object and introduce corresponding terminology. We then discuss the difficulty of achieving accessibility while guaranteeing flexibility in cobot handling tasks, followed by how they are both achieved by RTCoHand framework. Finally, we discuss the challenges that the cobot needs to address and propose corresponding solutions.

\subsection{Real-time cobot handling task and terminology}
\label{sec:rtcohand_terminology}

\begin{figure}[h]
    \centering
    \includegraphics[width=\linewidth]{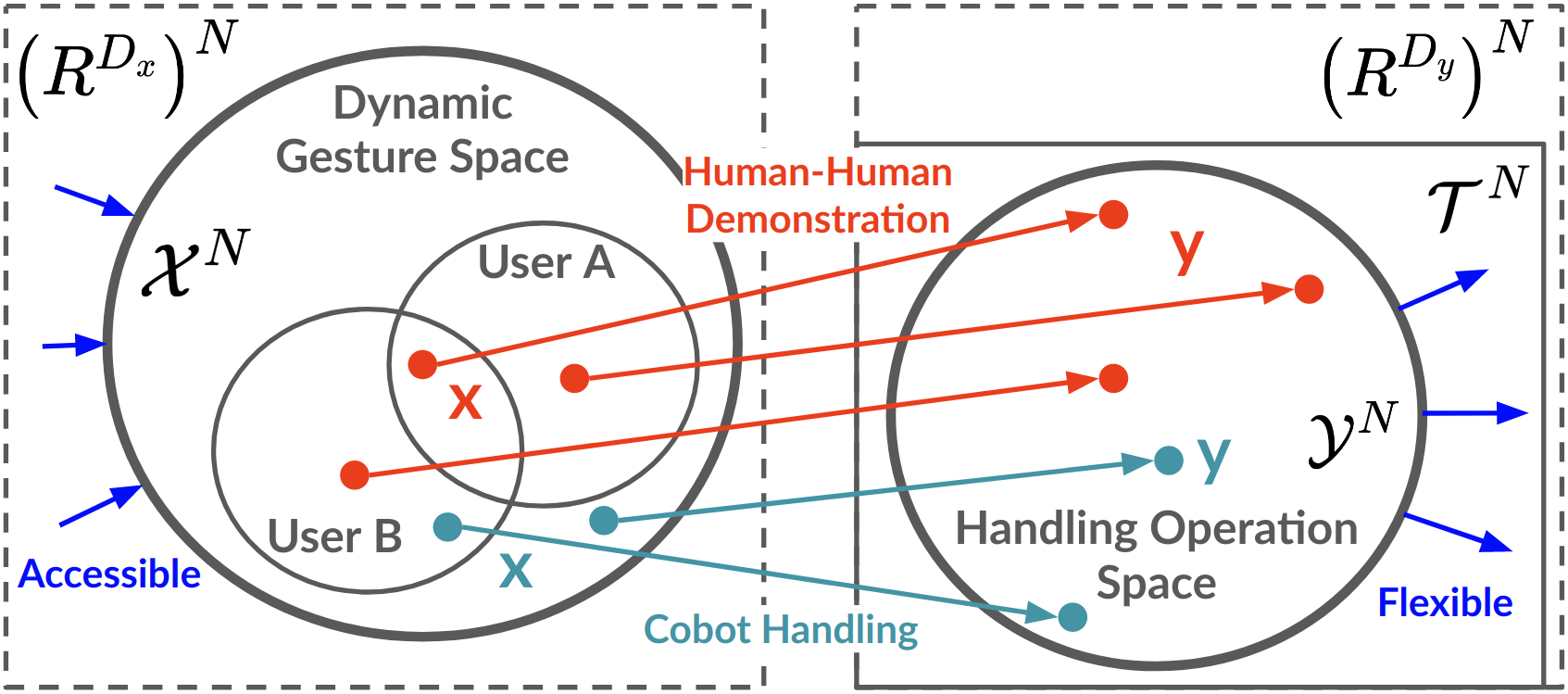}
    \caption{Illustration of abstract spaces in cobot handling tasks. To the left, we have the dynamic gesture space $\cX^N$ as a subspace of $D_x$-dimensional sequences. To the right, we have the handling operation space $\cY^N$ as a subspace of the Cartesian velocity sequence space $\cT^N$, which is a subspace of $D_y$-dimensional sequences. Different users (e.g., user A and user B) are allowed to customize and perform dynamic gestures from different regions (smallest circles) of $\cX^N$ for maximal comfort. We learn a cobot policy from human-human collaboration data (e.g., $(\bx,\by)$ pairs in red) and deploy it in real-time tasks (e.g., $(\bx, \by)$ pairs in green). Finally, a cobot handling task is more accessible if it only needs a small dynamic gesture space (blue suppression arrows on the left) to achieve a wide (flexible) range of possible operations (blue extention arrows on the right).}
    \label{fig:abstract_space_task}
\end{figure}

In the proposed RTCoHand framework, users control the cobot to handle an object by dynamic gestures. In this paper, we focus on only the right hand for simplicity, while our framework can be extended to using both hands directly. We first define dynamic gestures as continuous hand maneuvers involving change of hand position, hand orientation, and finger joint angles, denoted by $\bx\defeq \left(x^{(1)}, x^{(2)}, \dots, x^{(N)}\right)\in\cX^N$. Each $x^{(t)}\in \cX$ encodes the hand gesture at time $t$ and $\cX\subset\RR^{D_x}$. $\cX^N \defeq \cX\times\cX\dots\times\cX$ is the dynamic gesture space. Now, we define handling operation as a sequence of $6\mathrm{D}$ rigid body transformation velocities defined in Cartesian space, or Cartesian velocities. We denote the Cartesian velocity at each timestep as $y\in\cY$ and $\cY\subset\cT$, where $\cT\subset\RR^{D_y}$ is the space all possible Cartesian velocities (i.e., $D_y=6$). We then denote handling operation as $\by\defeq \left(y^{(1)}, y^{(2)}, \dots, y^{(N)}\right)\in\cY^N$ where $\cY^N \defeq \cY \times \cY\dots\times\cY$ is the operation space.
In this paper, we also refer to $\by$ as \textit{object motion} as situation fits.
Here, $N$ refers to the duration of any continuous cobot handling session, and can vary as needed. To command each desired operation $\by$ on the object, users perform a unique dynamic gesture $\bx$. To ease analysis, we assume both sequences to have the same length. In practice, the generated handling operation can be interpolated or sub-sampled if denser or sparser control is desired. Also, to practically implement this task, we assume both dynamic gesture space $\cX^N$ and handling operation space $\cY^N$ to be bounded. In this work, we propose \textit{ad hoc} constraints to bound $\cX^N$ and $\cY^N$ (to be discussed in section \ref{sec:collect_user_policies}), and leave systematic approaches as future work. See \cref{fig:abstract_space_task} for an illustration of the aforementioned spaces.

On the cobot side, when observing human command $\bx$, the cobot performs operation $\by$ according to a \textit{handling policy} $\pi_\theta:\cX^N\mapsto\cY^N$ parameterized by $\theta$. This policy should achieve human-level assistance as if a human helper were performing operations that match the users' expectations. In other words, the goal of cobot handling is to find a handling policy $\pi_\theta$ that solves the following minimization
\begin{equation}\label{eq:cobot_goal}
\minimizewrt{\theta} \Expect{(\bx,\by)\sim D_\mathsf{train}}{\norm{\pi_\theta(\bx)-\by}}.
\end{equation}
where $D_\mathsf{train}$ is a human-labeled dataset containing dynamic gestures and desired cobot handling operations and $\norm{\cdot}$ is the norm function. For fluent collaboration, we add a real-time requirement that the cobot computes and applies handling operation $\by$ at the same time when users perform commands $\bx$. See \cref{fig:task_diagram} for an illustration of cobot handling task.

% \ruic{discuss the definition of a single motion as similar transformations in a time range}

% The cobot always reacts to human commands at every time step which leads to real-time human-robot collaboration.

\subsection{Difficulty of accessibility under flexibility}\label{sec:difficulty_task}

Under definitions in the preceding section, \textbf{flexibility} of cobot handling tasks can be defined as proportion of all transformation sequences $\cT^N$ that is covered by possible operations $\cY^N$. \textbf{Accessibility} is then decided by the minimum size of dynamic gesture space $\cX^N$ such that we can find a valid handling policy $\pi_\theta$ satisfying $\forall \by\in\cY^N$, $\exists\bx\in\cX^N\Rightarrow \pi_\theta(\bx)=\by$. The smaller such size is, the less dynamic gestures users need to memorize and hence the better accessibility. Control using dynamic gestures is easy (accessible) when the target actions are simple (inflexible). For example, people have been using dynamic gestures to guide others to park their cars, where target actions might only include ``left'', ``right'', ``proceed'', and ``stop''. However, in cobot handling tasks, the target actions are general Cartesian velocities which lead to exponentially more possible decisions. They are significantly harder for users to match with dynamic gestures. One solution is to develop a ``manual'' that specifies dynamic gesture $\bx$ for all possible handling operations $\by$ and ask users to memorize. This approach indeed guarantees the flexibility of operations, but the designed dynamic gestures might be physically demanding for some users if they have limited strength and joint flexibility. In short, achieving accessibility is particularly hard because cobot handling operations are flexible while different users have varying physical capabilities. See \cref{fig:abstract_space_task} for an illustration of accessibility and flexibility under the definition of both dynamic gesture and handling operation spaces.

\begin{figure}[h]
    \centering
    \includegraphics[width=\linewidth]{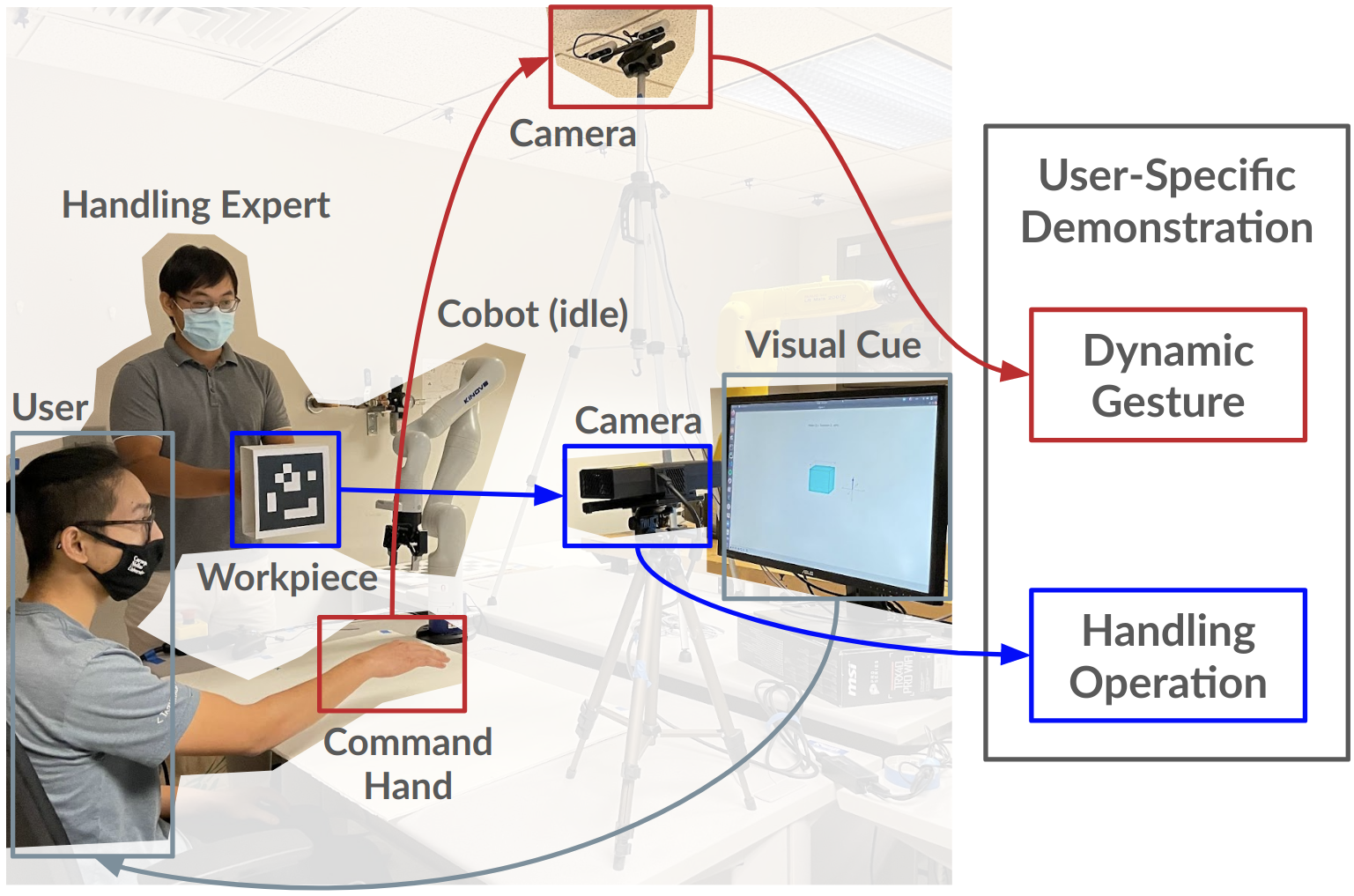}
    \caption{Human-human collaboration as demonstration for cobot handling tasks. In contrast to the test time scenario shown in \cref{fig:task_diagram}, the user performs dynamic gestures according to designated handling operations indicated by the visual cue. The expected real-time handling operations are labeled by a handling expert. The expert maneuvers an object that can be tracked using a camera. Both dynamic gestures and handling operations are stored in user-specific databases. Note that the cobot is idle in this process.}
    \label{fig:rtcohand_info}
\end{figure}

\subsection{RTCoHand framework in user perspective}\label{sec:framework_in_user}

To achieve maximum accessibility while enabling flexible handling operations, we propose Real-Time Collaborative Robot Handling (RTCoHand) framework. The key feature of this framework is \textit{policy customization}
% which allows users to customize and perform comfortable dynamic gestures $\bx$ for desired object motions $\by$. 
% Next, we introduce each of these features and discuss how they combine to boost accessibility while preserving all the flexibility we desire in cobot handling tasks. 
% \paragraph*{\textbf{Policy customization}}
which reduces users' physical burden by allowing them to customize comfortable dynamic gestures $\bx$ for desired handling operations $\by$ instead of following a pre-designed one. This is motivated by the fact that the same dynamic gesture can appear natural and intuitive for certain users but not for others due to physical limitations. For example, rotating the wrist without moving the arm might only appear easy to users with high wrist flexibilities. As such, it is impractical to design a common dynamic gestures set which accommodates all different user capabilities. Instead, we allow each user to use only the dynamic gestures that appear easy for that user. To obtain the corresponding cobot policy $\pi_\theta$, we can learn from human-human collaboration where the user demonstrates the handling task (in his/her customized way) with another human \textit{handling expert} acting in the role of cobot in real-time (see \cref{fig:rtcohand_info}). The resulting data would constitute the dataset $D_\mathsf{train}$ as in \eqref{eq:cobot_goal}. Detailed setup for collecting demonstration data will be covered in section \ref{sec:collect_user_policies}. As a summary, policy customization encourages users to exploit different portions of dynamic gesture space $\cX^N$ for maximal comfort (see \cref{fig:abstract_space_task}).

\subsection{RTCoHand framework in cobot perspective}\label{sec:cobot_goal}

In previous sections, we discuss how RTCoHand framework specifies an accessible and flexible cobot handling task for human users. In this section, we focus on the cobot  and discuss practical challenges we need to solve. For each challenge, we propose solutions in high-level descriptions and intuitions. We finally reach a mathematical formulation of the goal of cobot handling tasks.

% \paragraph*{\textbf{Challenge of solving inverse of command policies}}

% To achieve the cobot handling goal given by Eq.~\eqref{eq:cobot_goal}, a straightforward solution would be setting the cobot handling policy to the inverse of user command policy, or $\pi_\theta=\xi^{-1}$. However, this would require $\xi$ to be fully known which is impractical when we allow users to specify their own $\xi$. What is steadily available instead are the user commands $\bx=\xi(\by)$ corresponding to a finite range of handling operations $\by$. As such, in this paper, instead of exactly solving $\xi$ and its inverse, we propose to directly approximate a $\pi_\theta\approx\xi^{-1}$ using the demonstration $D_{\mathsf{train}}=\{(\bx_{\mathsf{train}},\by_{\mathsf{train}})\}$ we collect from users when they are customizing $\xi$. This can be done via empirical risk minimization (ERM) with the loss being the difference between cobot-inferred handling operations $\hat{\by}=\pi_\theta(\bx)$ and true user commands $\by=\xi^{-1}(\bx)$, formulated as
% \begin{equation}\label{optim:cobot_goal_1}
%     \minimizewrt{\theta} \Expect{(\bx,\by)\sim D_{\mathsf{train}}}{\cL\left( \pi_\theta(\bx), \by \right)}
% \end{equation}
% where $\cL$ is an appropriate loss function.

\paragraph*{\textbf{Challenge of adapting to different or new users}}

\begin{figure*}[t]
    \centering
    \begin{subfigure}[b]{0.45\linewidth}
        \centering
        \begin{tikzpicture}[node distance=1.2cm, thin, -latex, bend angle=45, auto]

    \colorlet{TrainColor}{OliveGreen}
    \colorlet{TestColor}{Blue}

    \tikzstyle{obs} = [circle, text centered, minimum size=1.0cm, draw=black, fill=black!10, semithick]
    \tikzstyle{latent} = [circle, text centered, minimum size=1.0cm, draw=black, fill=black!0, semithick]
    \tikzstyle{label} = [rectangle, minimum width=0.1cm, minimum height=0.1cm, text centered]

    \tikzstyle{entity} = [rectangle, rounded corners, minimum width=0.1cm, minimum height=0.6cm, text centered]
    \tikzstyle{entity_mid} = [rectangle, rounded corners, minimum width=0.1cm, minimum height=0.6cm, text centered]

    \tikzstyle{arrow} = [semithick,->,-latex]
    \tikzstyle{arrow_train} = [semithick,->,-latex,draw=TrainColor]
    \tikzstyle{arrow_test} = [semithick,->,-latex,draw=TestColor]

    \node [entity] (user) {$\mathsf{User}$};

    % train
    \node [latent] (x_train) [at=(user.west |- user.east), xshift=-1.75cm] {$\bx_{\mathsf{train}}$};
    \node [entity, align=center, font=\footnotesize] (human_expert) [below of=x_train] {$\mathsf{Human}$\\$\mathsf{Expert}$};
    \node [latent] (y_train) [below of=human_expert] {$\by_{\mathsf{train}}$};

    \draw [arrow] (user) to (x_train);
    \draw [arrow] (x_train) to (human_expert);
    \draw [arrow] (human_expert) to (y_train);

    % middle
    \node [entity_mid, font=\footnotesize, text=TrainColor] (learn) [at=(human_expert -| user.south), yshift=0.5cm] {$\mathsf{Learn}$};
    \node [entity_mid, font=\footnotesize, text=TestColor] (dep) [at=(human_expert -| user.south), yshift=-0.5cm] {$\mathsf{Context}$};

    % test
    \node [latent] (x_test) [at=(user.east |- user.east), xshift=1.75cm] {$\bx_{\mathsf{test}}$};
    \node [entity, align=center, font=\footnotesize] (cobot_policy) [below of=x_test] {$\mathsf{Cobot}$\\$\mathsf{Policy}$};
    \node [latent] (y_test) [below of=cobot_policy] {$\by_{\mathsf{test}}$};

    \draw [arrow] (user) to (x_test);
    \draw [arrow] (x_test) to (cobot_policy);
    \draw [arrow] (cobot_policy) to (y_test);

    \draw [arrow_train] (x_train) to (learn.west);
    \draw [arrow_train] (y_train) to (learn.west);
    \draw [arrow_train] (learn.east) to (cobot_policy.west);

    \draw [arrow_test] (x_train) to (dep.west);
    \draw [arrow_test] (y_train) to (dep.west);
    \draw [arrow_test] (dep.east) to (y_test.west);

    % \draw [arrow_test] (x_train.east) to[out=0,in=150] (cobot_policy);
    % \draw [arrow_test] (y_train.east) to[out=0,in=-150] (cobot_policy);

    \begin{pgfonlayer}{background}
        \filldraw [line width=3mm,join=round,black!20]
            (x_train.north  -| x_train.east) rectangle (y_train.south  -| y_train.west)
            (x_test.north  -| x_test.east) rectangle (y_test.south  -| y_test.west);
    \end{pgfonlayer}

    \node [label, font=\footnotesize, align=center] (pc) [above of=x_train, yshift=-0.1cm] {$\mathsf{Policy~Customization}$\\$\mathsf{(training)}$};
    \node [label, font=\footnotesize, align=center] (deployment) [above of=x_test, yshift=-0.1cm] {$\mathsf{Deployment}$\\$\mathsf{(testing)}$};

\end{tikzpicture}
        \caption{Training and testing phases of RTCoHand.}
    \label{fig:train_test_procedure}
    \end{subfigure}
    \hfill
    \begin{subfigure}[b]{0.54\linewidth}
        \centering
        \begin{tikzpicture}[node distance=1.4cm, thin, -latex, bend angle=45, auto]

    \tikzstyle{obs} = [circle, text centered, minimum size=1.0cm, draw=black, fill=black!10, semithick]
    \tikzstyle{latent} = [circle, text centered, minimum size=1.0cm, draw=black, fill=black!0, semithick]
    \tikzstyle{label} = [rectangle, minimum width=0.1cm, minimum height=0.1cm, text centered]

    \tikzstyle{arrow} = [semithick,->,-latex]
    \tikzstyle{arrow_bg} = [semithick,->,-latex,draw=black!100]
    
    % y target
    \node [obs] (yT1) {$y_T$};
    \node [obs] (yT2) [right of=yT1] {$y_T$};
    \node [label] (dots_y) [right of=yT2] {$\dots$};
    \node [obs] (yTN) [right of=dots_y] {$y_T$};
    \draw [arrow] (yT1) to (yT2);
    \draw [arrow] (yT2) to (dots_y);
    \draw [arrow] (dots_y) to (yTN);

    \node [latent] (z) [below of=yT2, xshift=0.8cm, yshift=0.3cm] {$z$};

    % x target
    \node [obs] (xT1) [above of=yT1, yshift=0.5cm] {$x_T$};
    \node [obs] (xT2) [above of=yT2, yshift=0.5cm] {$x_T$};
    \node [label] (dots_x) [above of=dots_y, yshift=0.5cm] {$\dots$};
    \node [obs] (xTN) [above of=yTN, yshift=0.5cm] {$x_T$};

    \draw [arrow] (xT1) to (yT1);
    \draw [arrow] (xT2) to (yT2);
    \draw [arrow] (xTN) to (yTN);

    \draw [arrow] (z) to (yT1.south);
    \draw [arrow] (z) to (yT2.south);
    \draw [arrow] (z) to (yT2.south -| dots_y.south);
    \draw [arrow] (z) to (yTN.south);

    \node [label] (t1) [above of=xT1, yshift=-0.6cm] {$t=1$};
    \node [label] (t2) [above of=xT2, yshift=-0.6cm] {$t=2$};
    \node [label] (tN) [above of=xTN, yshift=-0.6cm] {$t=N_T$};

    % context
    \node [obs] (xC) [left of=xT1, xshift=-0.2cm, yshift=-0.5cm] {$x_C$};
    \node [obs] (yC) [left of=yT1, xshift=-0.2cm] {$y_C$};

    \draw [join=round,black,rounded corners,semithick]
            ([xshift=-0.2cm, yshift=0.2cm] xC.west |- xC.north) rectangle ([xshift=0.2cm, yshift=-0.4cm] yC.south  -| yC.east);
    \node [label] (NC) [below of=yC, xshift=0.4cm, yshift=0.7cm] {$N_C$};

    % context to target
    \draw [arrow_bg] (xC) to[out=-5,in=135] (yT1.north west);
    \draw [arrow_bg] (yC) to[out=35,in=135] (yT1.north west);

    \draw [arrow_bg] (xC) to[out=-5,in=140] (yT2.north west);
    \draw [arrow_bg] (yC) to[out=35,in=140] (yT2.north west);

    \draw [arrow_bg] (xC) to[out=-3,in=160] (yTN.north west);
    \draw [arrow_bg] (yC) to[out=35,in=160] (yTN.north west);

    \node [label] (context_label) [above of=xC, yshift=-0.4cm] {$\mathsf{Context}$};
    \node [label] (target_label) [above of=xT2, yshift=-0.2cm, xshift=0.8cm] {$\mathsf{Target}$};

    % \begin{pgfonlayer}{background}
    %     \filldraw [line width=3mm,join=round,black!10]
    %         (f_enc_1.north  -| f_enc_1.east) rectangle (f_enc_3.south  -| f_enc_3.west);
    % \end{pgfonlayer}

\end{tikzpicture}
        \caption{Graphical model of cobot policy (CCHP).}
    \label{fig:pgm}
    \end{subfigure}
    \caption{(a) shows a computation diagram of RTCoHand framework. During policy customization (left), we collect $D_{\mathsf{train}}=\{(\bx_{\mathsf{train}},\by_{\mathsf{train}})\}$ for learning the cobot policy (green arrows). During deployment (right), the cobot policy generates handling operations $\by_\mathsf{test}$ for new user commands $\bx_\mathsf{test}$ conditioned on training data (blue arrows). (b) is a graphical model of CCHP that represents the probabilistic view of cobot policy. The policy encodes the randomness of cobot handling in a latent variable $z$. It also utilizes user-specific context $(\bx_C,\by_C)$ to support the intepretation of new user commands $\bx_T$ in real time. The conditional dependency between $y_T$ at adjacent time steps encourages smooth cobot actions. $N_C$ and $N_T$ are lengths of context and target respectively.}
    \label{fig:procedure_pgm}
\end{figure*}
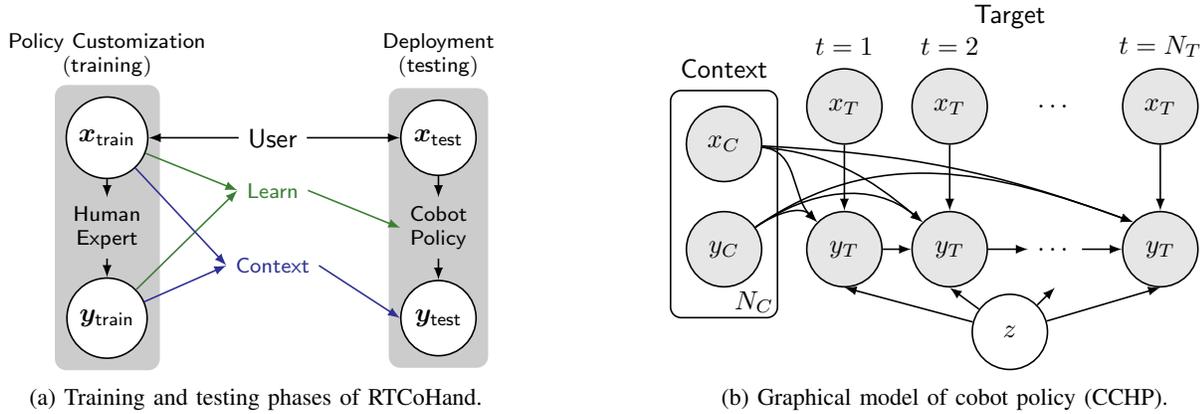

With policy customization, the dynamic gestures $\bx$ for each handling operation $\by$ from different users can vary significantly. Such diversity would also increase when new users appear. One naive solution is to train a unique handling policy $\pi_\theta$ for each existing or new user. However, this can be expensive and unscalable especially in industrial scenarios, where new workers should be quickly prepared for existing tasks. Hence, we need a single handling policy $\pi_\theta$ that can both adapt to different existing users and new users without having to repeat the training process. To achieve this, we assume access to some prior knowledge about users and introduce explicit dependence of $\pi_\theta$ on the prior. The prior is essentially some annotated command-operation pairs $\{(\bx_C,\by_C)\}$, referred to as $\textit{context}$. Then, we refer to new user commands and true desired operations as $\textit{target}$, denote as $\{(\bx_T,\by_T)\}$. In practice, we simply use the policy customization data $D_{\mathsf{train}}$ as context, while the target would be the data $D_{\mathsf{test}}=\{(\bx_\mathsf{test},\by_\mathsf{test})\}$ in actual deployment. In this spirit, we update the goal of cobot handling \eqref{eq:cobot_goal} with a conditional form as
\begin{equation}\label{optim:cobot_goal_2}
    \minimizewrt{\theta} \Expect{(\bx_T,\by_T)\sim D_{\mathsf{test}}}{\norm{\pi_\theta(\bx_T\mid \bx_C,\by_C) - \by_T}}
\end{equation}
where $(\bx_C,\by_C)\equiv D_\mathsf{train}$. All $(\bx,\by)$ pairs are labeled with the human demonstrator ID, so that we can match the context with user during testing. In this way, $\pi_\theta$ no longer needs to fully encode each user's preferences, since it can draw insights from user-specific database to aid the inference of new handling operations. Now, notice that \eqref{optim:cobot_goal_2} represents a testing (deployment) time objective and cannot be directly solved, because the true desired operation $\by_T$ is unavailable during training. Hence, at training time, we split the demonstration $D_{\mathsf{train}}$ into context and target to simulate the testing scenario, yielding a practical training time objective as follows
\begin{equation}\label{optim:cobot_goal_2_train}
    \minimizewrt{\theta} \Expect{(\bx_C,\by_C,\bx_T,\by_T)\sim D_{\mathsf{train}}}{\norm{\pi_\theta(\bx_T\mid \bx_C,\by_C) - \by_T}}.
\end{equation}
See \cref{fig:train_test_procedure} for an illustration of the training and testing phases of RTCoHand framework. Our idea of solving adaptation via conditional prediction is partially inspired by the line of research on NPs \cite{cnp,np,anp} and conceptually similar to few-shot learning, where the target data is compared to observed data in some feature space \cite{siamese, one_shot_with_mem_augment, vinyals_matching_2017, bartunov_fast_2017}. For more detailed analysis of such connection, we refer readers to \cite{cnp,np,anp}.

\paragraph*{\textbf{Challenge of modeling human motion uncertainties}}

The next challenge we identify is the uncertainty of user commands $\bx$. We argue that during policy customization, users develop dynamic gestures in high-level intuitions such as ``rotate the wrist'' and ``move the forearm'', instead of hand trajectories with exact numerical values. As such, the actual dynamic gestures $\bx$ carried out by users for the same desired operation $\by$ might vary among multiple trails. As a result, the desired operation $\by$ for a given command $\bx$ might also be uncertain. Hence, to be more general, we promote the handling policy $\pi_\theta$ to a stochastic version $\Pi_\theta$ where $\Pi_\theta(\bx_T)\sim p_\theta(\by_T\mid\bx_T,\bx_C,\by_C)$. $p_\theta$ is a conditional probability modeling the human-human demonstration. With that in hand, we arrive at our final formulation of the cobot handling goal as learning a conditional distribution:
\begin{equation}\label{optim:cobot_goal_3}
    \maximizewrt{\theta} \Expect{(\bx_C,\by_C,\bx_T,\by_T)\sim D_\mathsf{train}}{\log p_\theta(\by_T\mid\bx_T,\bx_C,\by_C)}.
\end{equation}
% We provide further analysis of the high-level objective \eqref{optim:cobot_goal_3} and a corresponding learning algorithm in section \ref{sec:model_theory}.

% \ruic{Users would not exactly repeat the same $\bx$ every time for similar desired $\by$, so the mapping between command and operation can be described using a probability distribution instead of a deterministic function.}

\paragraph*{\textbf{Challenge of robustness against noisy input for safety}}

In previous challenges, we focus on how the cobot can generate reasonable handling operations considering user variation and uncertainty. To deploy the cobot in real scenarios, we also need to ensure safety. In cobot handling tasks, we consider cobot motions to be safe if the trajectory is consistent and smooth, without abrupt movements or large accelerations even when human input is noisy. Naturally, all handling operations collected in prior should satisfy this safe requirement. As such, we just need to enable the cobot to capture the smoothness of $\by_T$ during training and ensure the same during testing. To achieve this, we propose a transition probability between adjacent handling operations $y_T^{(t-1)}$ and $y_T^{(t)}$ to encode the continuity in $\by_T$. Here, we provide this temporal dependency as an intuition and leave corresponding analytical representation to the next section.

\section{Conditional Collaborative Handling Process}
\label{sec:model_theory}

In this section, we provide an probabilistic view of cobot handling policy $\Pi_\theta$ and formally propose {\em conditional collaborative handling process} (CCHP) to model the underlying generative process. Then, we derive the learning objective of CCHP for solving our cobot handling goal \eqref{optim:cobot_goal_3}.

\subsection{Probabilistic Perspective of Cobot Handling}\label{sec:cchp_intro}

We desire a stochastic cobot handling policy $\Pi_\theta$ which samples from a probability distribution conditioning on user-specific prior, i.e., $\Pi_\theta(\bx_T)\sim p_\theta(\by_T\mid\bx_T,\bx_C,\by_C)$. To model this distribution, we propose {\em conditional collaborative handling process} (CCHP). See \cref{fig:pgm} for a graphical representation. The CCHP incorporates three key features, each attending to a practical challenge (see section \ref{sec:cobot_goal}). First, we incorporate explicit dependence of $\Pi_\theta$ on a user-specific context $(\bx_C,\by_C)$ to adapt to different users. Then, we introduce a latent variable $z\in\RR^{D_z}$ to capture the underlying randomness of handling operations $\by_T$. Intuitively, $z$ encodes a wide range of characteristics of handling operations, e.g., how dynamic gesture patterns map to those in handling operations, overall scale of cobot movements, and amount of variety (uncertainty). Finally, we setup temporal dependency in $\by_T$ to capture the smoothness of handling operations to ensure robustness and safety under noisy input. Hence, we arrive at the following generative process
\begin{align}\label{eq:generative_process}
    & p_\theta(\by_T\mid \bx_T,\bx_C,\by_C)\nonumber \\
    \defeq \int & p_\theta(\by_T\mid \bx_T,\bx_C,\by_C,z) p(z) dz. \nonumber \\
    = \int & \prod_{t=1}^{N_T} p_\theta(y_T^{(t)} \mid \by_T^{(1:t-1)}, x_T^{(t)},\bx_C,\by_C,z) p(z) dz.
\end{align}
where $\by_T^{(1:t-1)}\defeq (y_T^{(1)},y_T^{(2)},\dots,y_T^{(t-1)})$. As long as the empirical data we learn with (i.e., $D_\mathsf{train}$) contains only safe and smooth handling operations, the temporal dependency of $\by_T$ would prevent the current action $y_T^{(t)}$ from heavily deviating from the past trajectory $\by_T^{(1:t-1)}$. This in turn improves the safety and smoothness during actual real-time cobot handling.

% Formulate problem, describe using PGM, discuss the need ot context. Introduce terminology. \ruic{mention time dependency}

\subsection{Learning and inference of CCHP}\label{sec:learn_infer_CCHP}

To learn the distribution \eqref{eq:generative_process} from data, we approximate the posterior of $z$ using a variational distribution $q_\phi(z\mid \bx_T,\by_T,\bx_C,\by_C)$ and minimize its Kullback–Leibler (KL) divergence with the true posterior, given by
\begin{equation}\label{eq:KL}
    \minimizewrt{\phi} D_{\mathsf{KL}}(q_\phi(z\mid \bx_T,\by_T,\bx_C,\by_C) \mid\mid p(z\mid \bx_T,\by_T,\bx_C,\by_C))
\end{equation}
where $\phi$ parameterizes the varational posterior $q$. It can be shown that solving \eqref{eq:KL} is equivalent to maximizing the following evidence lower bound (ELBO):
\begin{align}
    & \log p(\by_T\mid \bx_T,\bx_C,\by_C) \geq \mathsf{ELBO}\label{eq:elbo_1}
\end{align}
where $\mathsf{ELBO}$ is given by
\begin{align}
    \mathsf{ELBO} &\defeq \Expect{z\sim q_\phi(z\mid \bx_T,\by_T,\bx_C,\by_C)}{\log p_\theta(\by_T\mid \bx_T,\bx_C,\by_C,z)} \nonumber \\
    & - D_\mathsf{KL}(q_\phi(z\mid \bx_T,\by_T,\bx_C,\by_C) \mid\mid p(z)). \label{eq:ELBO_def}
\end{align}
See Appendix~\ref{append:KL_ELBO} for a detailed derivation of \eqref{eq:elbo_1} and \eqref{eq:ELBO_def}. In the ELBO \eqref{eq:ELBO_def}, the first term is the expected log-likelihood of observed $\by_T$. Maximizing this term would encourage CCHP to explain the observation with its generative process. The second term is a regularization term that prevents the posterior from deviating too much from the prior during inference process. Now, note that to optimize \eqref{eq:ELBO_def}, we need to know the prior $p(z)$, which is usually either intractable or assumed to be known (e.g., standard Gaussian). In this work, we follow \cite{attitude_2017} to approximate it using the variational posterior $q_\phi(z\mid \bx_C,\by_C)$ conditioned on context data only. Instead of using an uninformed prior as in the case of variational autoencoders \cite{kingma2014autoencoding}, we are extracting useful information about each user from a personalized database. The KL term in \eqref{eq:ELBO_def} then essentially keeps the posterior distribution consistent within each user, conditioned on either context (past interaction) or target (new interaction). This can be more clearly seen in the following ELBO form:
\begin{align}
    \mathsf{ELBO}=&\sum_{t=1}^{N_T} \Expect{z\sim q_{\phi\mid T}}{\log p_\theta(y_T^{(t)}\mid \by_T^{(1:t-1)}, x_T^{(t)},\bx_C,\by_C,z)} \nonumber \\
     - D_\mathsf{KL} & (q_\phi(z\mid \bx_T,\by_T,\bx_C,\by_C) \mid\mid q_\phi(z\mid \bx_C,\by_C)). \label{eq:elbo_2}
\end{align}
where $q_{\phi\mid T}$ abbreviates $q_\phi(z\mid \bx_T,\by_T,\bx_C,\by_C)$. Notice that in \eqref{eq:elbo_2}, we also incorporate the temporal structure of $\by_T$ by expanding $p_\theta(\by_T\mid \bx_T,\bx_C,\by_C,z)$ for each time step.

To practically solve \eqref{eq:elbo_2}, we assume that both the generative model $p$ and the approximate inference model $q$ are Gaussian distributions that can be parameterized by learnable functions, e.g., artificial neural networks. Specifically, we assume a parameterized Gaussian posterior $q_\phi(z\mid\bx_*,\by_*)$ as $z\sim\cN\left(\mu_\phi,\Sigma_\phi\right)$. We assume $\mu_\phi, \Sigma_\phi = F_\mathsf{enc}(\bx_*,\by_*\mid \phi)$ where $F_\mathsf{enc}$ is a nonlinear \textit{encoder} function, represented by a neural network with parameter $\phi$. When $(\bx_*,\by_*)$ involves target data, we refer to the resulting distribution \textit{latent posterior}, or $q_{\phi\mid T}$ as in \eqref{eq:elbo_2}. When $(\bx_*,\by_*)$ contains only context data, we have the \textit{approximate latent prior}, or $q_{\phi\mid C}\defeq q_\phi(z\mid \bx_C,\by_C)$. Similarly, we assume the generative process $p_\theta(y_T^{(t)}\mid \by_T^{(1:t-1)}, x_T^{(t)},\bx_C,\by_C,z)$ as $y_T^{(t)}\sim\cN\left(\mu_\theta^{(t)},\Sigma_\theta^{(t)}\right)$ where $\mu_\theta^{(t)}$ and $\Sigma_\theta^{(t)}$ are generated by a neural network \textit{decoder} $F_\mathsf{dec}(\by_T^{(1:t-1)}, x_T^{(t)},\bx_C,\by_C,z\mid \theta)$. Now, we can write the goal of cobot handling as the following practical form:
\begin{equation}
    \minimizewrt{\phi,\theta} \sum_{t=1}^{N_T} \Expect{z\sim q_{\phi\mid T}}{\log p(y_T^{(t)}\mid \mu_\theta^{(t)},\Sigma_\theta^{(t)})} - D_\mathsf{KL} (q_{\phi\mid T} \mid\mid q_{\phi\mid C}). \label{eq:elbo_3}
\end{equation}

The above problem can be solved using gradient-based methods due to neural network parameterization. By solving \eqref{eq:elbo_3}, we improve inference on the latent $z$ while learning the generative model $p_\theta(\by_T\mid \bx_T,\bx_C,\by_C,z)$ from offline data $D_\mathsf{train}$. During online deployment, we compute cobot action by sampling from the learned generative model $p_\theta$, where the latent $z$ is sampled from the approximate latent prior $q_{\phi\mid C}$ conditioned on the user-specific context. In the next section, we describe the network architecture of both encoder $q_\phi$ and decoder $p_\theta$.

% \subsection{[optional] Performance Guarantee}

\section{Neural Network Architecture of CCHP}\label{sec:arch}

\begin{figure*}[ht]
    \centering

    \begin{subfigure}[b]{\linewidth}
        \centering
        \input{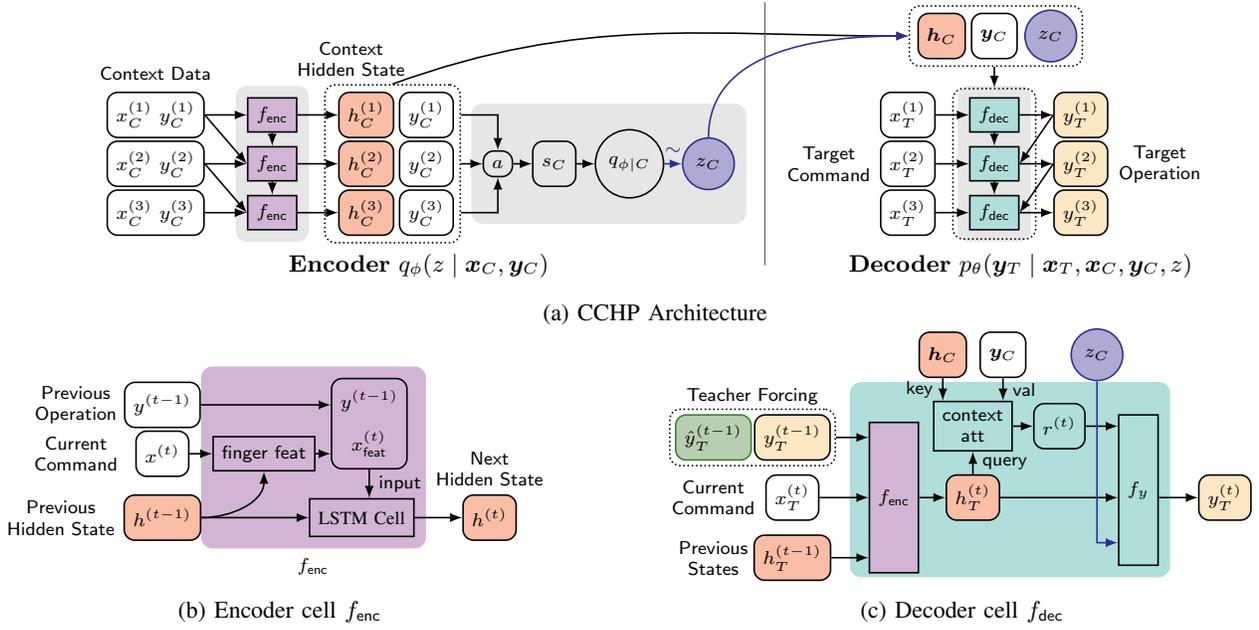}
        \caption{CCHP Architecture}
        \label{fig:CCHP_arch}
    \end{subfigure}\\

    \begin{subfigure}[b]{0.45\linewidth}
        \centering
        \colorlet{TrainColor}{OliveGreen}
\colorlet{TestColor}{Blue}
\colorlet{TrainColorFill}{TrainColor!30}
\colorlet{TestColorFill}{TestColor!30}

\colorlet{PredColor}{Dandelion}
\colorlet{PredColorFill}{PredColor!30}

\colorlet{HiddenColor}{Red!30}
\colorlet{zPriorColor}{TestColorFill}
\colorlet{zPostColor}{TrainColorFill}
\colorlet{FACColor}{Plum!30}
\colorlet{DecCellColor}{Emerald!30}

\begin{tikzpicture}[node distance=0.65cm, thin, -latex, bend angle=45, auto, font=\scriptsize]

    % normal var
    \tikzstyle{var} = [rectangle, rounded corners, minimum width=0.1cm, minimum height=0.6cm, text centered, draw=black, semithick]

    % hidden context
    \tikzstyle{var_hidden} = [rectangle, rounded corners, minimum width=0.1cm, minimum height=0.6cm, text centered, draw=black, fill=HiddenColor, semithick]

    % training var
    \tikzstyle{var_train} = [rectangle, rounded corners, minimum width=0.1cm, minimum height=0.6cm, text centered, draw=TrainColor, semithick, fill=TrainColorFill]

    % pred var
    \tikzstyle{var_pred} = [rectangle, rounded corners, minimum width=0.1cm, minimum height=0.6cm, text centered, draw=black, semithick, fill=PredColorFill]

    % select
    \tikzstyle{select} = [rectangle, rounded corners, text centered, draw=black,semithick, densely dotted]

    % dist
    \tikzstyle{dist} = [circle, text centered, draw=black,semithick]

    % z prior
    \tikzstyle{realis_prior} = [circle, text centered, draw=TestColor,semithick,fill=zPriorColor]

    % z post
    \tikzstyle{realis_post} = [circle, text centered, draw=TrainColor,semithick,fill=zPostColor]

    % Phi
    \tikzstyle{feat} = [rectangle, minimum width=0.4cm, minimum height=1.7cm,text centered, draw=black,semithick]

    % recurrent cells
    \tikzstyle{cell} = [rectangle, minimum width=0.1cm, minimum height=0.1cm, text centered, draw=black,semithick]

    % recurrent cells train
    \tikzstyle{cell_train} = [rectangle, minimum width=0.1cm, minimum height=0.1cm, text centered, draw=TrainColor,semithick, fill=TrainColorFill]

    % aggregate
    \tikzstyle{aggre} = [rectangle, rounded corners, minimum width=0.1cm, minimum height=0.1cm, text centered, draw=black,semithick]

    % label
    \tikzstyle{label} = [rectangle, minimum width=0.1cm, minimum height=0.1cm, text centered]

    % arrows
    \tikzstyle{arrow} = [semithick,->,-latex]
    \tikzstyle{arrow_train} = [semithick,->,-latex,draw=TrainColor]
    \tikzstyle{arrow_test} = [semithick,->,-latex,draw=TestColor]

    % encoder cell
    \begin{scope}

        % input
        % \node [var] (y_feat_tm1) {$y^{(t-1)}$};
        \node [var] (x_feat_t) {$x^{(t)}$};
        \node [cell] (FA) [right of=x_feat_t, xshift=0.7cm] {$\mathsf{finger~feat}$};
        \node [var_hidden] (h_tm1) [below of=x_feat_t, yshift=-0.2cm] {$h^{(t-1)}$};
        \node [var] (y_feat_tm1) [above of=x_feat_t, yshift=0.0cm] {$y^{(t-1)}$};
        % \node [var] (c_tm1) [below of=h_tm1] {$c^{(t-1)}$};

        % labels
        \node [label, align=center] (yt_1_label) [left of=y_feat_tm1, xshift=-0.5cm] {$\mathsf{Previous}$\\$\mathsf{Operation}$};
        \node [label, align=center] (xt_1_label) [left of=x_feat_t, xshift=-0.5cm] {$\mathsf{Current}$\\$\mathsf{Command}$};
        \node [label, align=center] (ht_1_label) [left of=h_tm1, xshift=-0.7cm] {$\mathsf{Previous}$\\$\mathsf{Hidden~State}$};

        \draw [arrow] (x_feat_t.east) -- (FA.west);
        \draw [arrow] (h_tm1.east) to [out=0,in=-90] (FA.south);
        \node [var, align=center, minimum height=1.2cm] (x_feat_att_t) [right of=FA, xshift=0.75cm, yshift=0.4cm] {$y^{(t-1)}$\\~\\$x^{(t)}_{\mathsf{feat}}$};
        \draw [arrow] (FA.east) -- (x_feat_att_t.west |- FA.east);
        \draw [arrow] (y_feat_tm1.east) -- (x_feat_att_t.west |- y_feat_tm1.east);
        
        \node [cell, minimum height=0.5cm] (lstm_cell) [right of=h_tm1, xshift=2.0cm] {LSTM Cell};
        \draw [arrow] (h_tm1.east) -- (h_tm1.east -| lstm_cell.west);
        % \draw [arrow] (c_tm1.east) -- (c_tm1.east -| lstm_cell.west);

        % \node [select, minimum height=1.65cm, minimum width=1.2cm, xshift=0.72cm, yshift=0.425cm] (lstm_input) [right of=FA] {};
        \draw [arrow] (x_feat_att_t.south) -- node[right] {$\mathsf{input}$} (lstm_cell.north -| x_feat_att_t.south);

        \node [var_hidden] (h_t) [right of=h_tm1, xshift=3.7cm] {$h^{(t)}$};
        \node [label, align=center] (ht_label) [above of=h_t] {$\mathsf{Next}$\\$\mathsf{Hidden~State}$};
        % \node [var] (c_t) [right of=c_tm1, xshift=3.5cm] {$c^{(t)}$};
        \draw [arrow] (h_t.west -| lstm_cell.east) -- (h_t.west);
        % \draw [arrow] (c_t.west -| lstm_cell.east) -- (c_t.west);

        % label
        \node [label] (encoder_cell_label) [below of=lstm_cell, yshift=-0.0cm, xshift=-0.65cm] {$f_\mathsf{enc}$};
        
    \end{scope}

    \begin{pgfonlayer}{background}
        \filldraw [line width=3mm,join=round,FACColor]
            (x_feat_att_t.north  -| lstm_cell.east) rectangle (lstm_cell.south  -| FA.west);
    \end{pgfonlayer}

\end{tikzpicture}
        \caption{Encoder cell $f_\mathsf{enc}$}
        \label{fig:enc_arch}
    \end{subfigure}%
    \begin{subfigure}[b]{0.55\linewidth}
        \centering
        \colorlet{TrainColor}{OliveGreen}
\colorlet{TestColor}{Blue}
\colorlet{TrainColorFill}{TrainColor!30}
\colorlet{TestColorFill}{TestColor!30}

\colorlet{PredColor}{Dandelion}
\colorlet{PredColorFill}{PredColor!30}

\colorlet{HiddenColor}{Red!30}
\colorlet{zPriorColor}{TestColorFill}
\colorlet{zPostColor}{TrainColorFill}
\colorlet{FACColor}{Plum!30}
\colorlet{DecCellColor}{Emerald!30}

\begin{tikzpicture}[node distance=0.65cm, thin, -latex, bend angle=45, auto, font=\scriptsize]

    % normal var
    \tikzstyle{var} = [rectangle, rounded corners, minimum width=0.1cm, minimum height=0.6cm, text centered, draw=black, semithick]

    % hidden context
    \tikzstyle{var_hidden} = [rectangle, rounded corners, minimum width=0.1cm, minimum height=0.6cm, text centered, draw=black, fill=HiddenColor, semithick]

    % training var
    \tikzstyle{var_train} = [rectangle, rounded corners, minimum width=0.1cm, minimum height=0.6cm, text centered, draw=TrainColor, semithick, fill=TrainColorFill]

    % pred var
    \tikzstyle{var_pred} = [rectangle, rounded corners, minimum width=0.1cm, minimum height=0.6cm, text centered, draw=black, semithick, fill=PredColorFill]

    % select
    \tikzstyle{select} = [rectangle, rounded corners, text centered, draw=black,semithick, densely dotted]

    % dist
    \tikzstyle{dist} = [circle, text centered, draw=black,semithick]

    % z prior
    \tikzstyle{realis_prior} = [circle, text centered, draw=TestColor,semithick,fill=zPriorColor]

    % z post
    \tikzstyle{realis_post} = [circle, text centered, draw=TrainColor,semithick,fill=zPostColor]

    % Phi
    \tikzstyle{feat} = [rectangle, minimum width=0.4cm, minimum height=1.7cm,text centered, draw=black,semithick]

    % recurrent cells
    \tikzstyle{cell} = [rectangle, minimum width=0.1cm, minimum height=0.1cm, text centered, draw=black,semithick]

    % recurrent cells train
    \tikzstyle{cell_train} = [rectangle, minimum width=0.1cm, minimum height=0.1cm, text centered, draw=TrainColor,semithick, fill=TrainColorFill]

    % aggregate
    \tikzstyle{aggre} = [rectangle, rounded corners, minimum width=0.1cm, minimum height=0.1cm, text centered, draw=black,semithick]

    % label
    \tikzstyle{label} = [rectangle, minimum width=0.1cm, minimum height=0.1cm, text centered]

    % arrows
    \tikzstyle{arrow} = [semithick,->,-latex]
    \tikzstyle{arrow_train} = [semithick,->,-latex,draw=TrainColor]
    \tikzstyle{arrow_test} = [semithick,->,-latex,draw=TestColor]

    % decoder cell
    \begin{scope}

        % input
        \node [var_train] (y_feat_T_tm1) {$\hat{y}^{(t-1)}_T$};
        \node [var_pred] (y_feat_pred_tm1) [right of=y_feat_T_tm1, xshift=0.4cm] {$y^{(t-1)}_T$};
        \node [select, minimum height=0.75cm, minimum width=2.2cm, xshift=-0.13cm] (tf_sel) [right of=y_feat_T_tm1] {};
        \node [label] (tf_label) [above of=tf_sel, yshift=-0.1cm] {$\mathsf{Teacher~Forcing}$};
        \node [var] (x_feat_T_t) [below of=y_feat_pred_tm1, xshift=0.0cm, yshift=-0.15cm] {$x^{(t)}_T$};
        \node [var_hidden] (h_tm1_dec) [below of=x_feat_T_t, yshift=-0.15cm] {$h^{(t-1)}_T$};
        % \node [var] (c_tm1_dec) [right of=h_tm1_dec, xshift=0.4cm] {$c^{(t-1)}$};
        % \node [select, minimum height=0.75cm, minimum width=2.2cm, xshift=-0.13cm] (f_dec_in_sel) [right of=h_tm1_dec] {};
        
        \node [label] (x_feat_T_t_label) [left of=x_feat_T_t, xshift=-0.4cm, align=center] {$\mathsf{Current}$\\$\mathsf{Command}$};
        \node [label] (h_t_1_label) [left of=h_tm1_dec, xshift=-0.4cm, align=center] {$\mathsf{Previous}$\\$\mathsf{States}$};

        % fac
        \node [feat, minimum height=2.0cm, fill=FACColor] (dec_fac) [right of=x_feat_T_t, xshift=0.7cm, yshift=0.0cm] {$f_\mathsf{enc}$};
        \draw [arrow] (tf_sel.east) -- (tf_sel.east -| dec_fac.west);
        \draw [arrow] (h_tm1_dec.east) -- (h_tm1_dec.east -| dec_fac.west);
        \draw [arrow] (x_feat_T_t.east) -- (x_feat_T_t.east -| dec_fac.west);

        % hiddens
        \node [var_hidden] (h_t_dec) [right of=dec_fac, xshift=0.4cm] {$h^{(t)}_T$};
        % \node [var] (c_t_dec) [below of=dec_fac, yshift=-0.76cm] {$c^{(t)}_T$};
        \draw [arrow] (h_t_dec.west -| dec_fac.east) -- (h_t_dec.west);
        % \draw [arrow] (dec_fac.south -| c_t_dec.north) -- (c_t_dec.north);

        % Context att
        \node [cell, align=center] (context_att) [above of=h_t_dec, xshift=0.0cm, yshift=0.3cm] {$\mathsf{context}$\\$\mathsf{att}$};
        \node [var_hidden] (hC) [above of=context_att, xshift=-0.4cm, yshift=0.3cm] {$\bh_C$};
        \node [var] (y_feat_C) [above of=context_att, xshift=0.4cm, yshift=0.3cm] {${\by}_C$};
        \draw [arrow] (hC.south) -- node[left]{$\mathsf{key}$} (hC.south |- context_att.north);
        \draw [arrow] (y_feat_C.south) -- node[right]{$\mathsf{val}$} (y_feat_C.south |- context_att.north);
        \draw [arrow] (h_t_dec.north) to node[right, xshift=0.0cm, yshift=0.0cm]{$\mathsf{query}$} (context_att);

        \node [var] (rt) [right of=context_att, xshift=0.5cm] {$r^{(t)}$};
        \draw [arrow] (context_att.east) -- (rt.west);
        
        % \node [cell_train] (latent_path) [right of=dec_fac, xshift=2.0cm, yshift=-0.5cm] {$\mathsf{latent~path}$};
        % \draw [arrow_train] (h_t_dec.east) to [out=0,in=180] (latent_path.west);

        % \node [realis_post] (z_post) [right of=latent_path, xshift=0.8cm] {$z_{T}$};
        % \draw [arrow_train] (latent_path.east) -- node[above] {$\textcolor{TrainColor}{\sim}$} (z_post.west);

        % \node [var_train] (y_feat_T_t) [below of=latent_path, yshift=-0.3cm] {$y^{(t)}_T$};
        % \draw [arrow_train] (y_feat_T_t.north) -- (latent_path.south);

        % output mlp
        \node [feat, minimum height=2.0cm] (dec_mlp) [right of=dec_fac, xshift=2.6cm, yshift=0.1cm] {$f_y$};
        \draw [arrow] (rt.east) -- (rt.east -| dec_mlp.west);
        % \draw [arrow_train] (z_post.east) -- (z_post.east -| dec_mlp.west);
        \draw [arrow] (h_t_dec.east) to [out=0,in=180] (dec_fac.east -| dec_mlp.west);

        \node [realis_prior] (z_C) [right of=y_feat_C, xshift=0.6cm] {$z_C$};
        \draw [arrow_test] (z_C.south) |- node[pos=0.25,name=5]{} ([yshift=-0.7cm]dec_mlp.west);
        % \draw [arrow] (d2) --  node[above=2pt] {no} ++(3,0) |- (p2);

        % output
        \node [var_pred] (y_pred_t) [right of=h_t_dec, xshift=2.7cm] {$y^{(t)}_T$};
        % \node [var_pred] (y_sig_t) [right of=z_post, xshift=1.55cm] {$\sigma^{(t)}_T$};

        \draw [arrow] (y_pred_t.west -| dec_mlp.east) -- (y_pred_t.west);
        % \draw [arrow] (y_sig_t.west -| dec_mlp.east) -- (y_sig_t.west);
        
        % label
        % \node [label] (decoder_cell_label) [above of=dec_mlp, yshift=1.1cm, xshift=-0.3cm] {$f_\mathsf{dec}$};

    \end{scope}

    % todo shade decoder cell, change ypred color
    \begin{pgfonlayer}{background}
        \filldraw [line width=3mm,join=round,DecCellColor]
            ([yshift=0.1cm]context_att.north  -| dec_mlp.east) rectangle ([xshift=-0.1cm] dec_mlp.south  -| dec_fac.west);
    \end{pgfonlayer}

    % separators
    % \draw [-, thick, gray] (-1.5cm, -2.4cm) -- (16cm, -2.4cm);
    % \draw [-, thick, gray] (6.5cm, -2.4cm) -- (6.5cm, -6.2cm);

\end{tikzpicture}
        \caption{Decoder cell $f_\mathsf{dec}$}
        \label{fig:dec_arch}
    \end{subfigure}%
    
    \caption{Computation diagram of the generative process of CCHP. In high-level diagram (a), the encoder (left) represent the posterior $q_\phi$ conditioned on context. ``$\sim$'' means sample operation. The decoder (right) represents the generative model $p_\theta(\by_T\mid \bx_T,\bx_C,\by_C,z)$. Both encoder and decoder have a recurrent structure. The structure of each individual cell is shown in (b) and (c), shaded with matching color ($f_\mathsf{enc}$ in purple and $f_\mathsf{dec}$ in cyan).}
    \label{fig:model_arc}
\end{figure*}

We now present a practical implementation of CCHP. See \cref{fig:model_arc} for a computation diagram. In following sections, we present in a top-down fashion, first capturing the overall structure as a probabilistic generative process and then going though fine-grained structures designed specifically for cobot handling tasks. Finally, we mention a robustness issue with training CCHP in practice and how it is resolved using a technique called \textit{teacher forcing}.

\subsection{High-level Structure}\label{sec:cchp_highlvl}

The overall structure of CCHP is shown in \cref{fig:CCHP_arch}. In the left segment, we implement the variational posterior $q_\phi$ as an encoder. The encoder takes any dynamic gestures $\bx$ and corresponding handling operations $\by$ as input and generates a Gaussian posterior of latent $z$. In the right segment, we implement the likelihood $p_\theta$ as an decoder. The decoder takes current dynamic gestures $\bx_T$ and sampled $z$ as input, and generates a Gaussian likelihood for the desired handling operation $\by_T$, conditioned on some context information.

Note that \cref{fig:model_arc} corresponds to the testing time where we sample $z$ from approximate prior $q_\phi(z\mid \bx_C,\by_C)$ in the generative process. In that case, we only feed context data to the encoder (see \cref{fig:CCHP_arch}). When necessary to infer the posterior, such as in case of solving \eqref{eq:elbo_3}, we can feed both context and target data to the encoder and get $q_\phi(z\mid \bx_T,\by_T,\bx_C,\by_C)$, while the network structure of $q_\phi$ remains the same. In the following section, we describe the encoder architecture in terms of a general form of input $(\bx,\by)$ for simplicity.
% In practice, such input can contain either only context or both context and target data.

\subsection{Encoder $q_\phi$}\label{sec:cchp_encoder}

A complete diagram of the encoder is shown in \cref{fig:CCHP_arch} (left). Given an input cobot handling trajectory $(\bx,\by)$ where $\bx\in\RR^{N\times D_x},\by\in\RR^{N\times D_y}$, the encoder first computes a hidden state $\bh$ that encodes the input dynamic gesture command at each time step. Here, $\bh\defeq (h^{(1)}, h^{(2)},\dots,h^{(N)})$ where $h^{(t)}\in\RR^{H}$ and $H$ is the hidden state size. Since we explicitly consider the temporal dependency in cobot handling operations (see Eq.~\eqref{eq:generative_process}), we want to consider the past operations $\by^{(1:t-1)}$ when trying to generate $y^{(t)}$ at current time step. Hence, when interpreting the current command $x^{(t)}$, i.e., computing $h^{(t)}$, we also feed $\by^{(1:t-1)}$ as input. This leads to the recurrent structure (the leftmost shaded area) we see in \cref{fig:CCHP_arch}. This structure concatenates several instances of the same function, referred to as \textit{encoder cell} ($f_\mathsf{enc}$), explained as follows.

\paragraph*{\textbf{Encoder Cell} $f_\mathsf{enc}$}

The encoder cell, as shown in \cref{fig:enc_arch}, interprets the current user command $x^{(t)}$ into a hidden state $h^{(t)}$ while considering the previous handling operation $y^{(t-1)}$ and previous hidden state $h^{(t-1)}$. This is achieved by the combination of a finger feature module and an long short-term memory (LSTM) cell. In the finger feature module, we first map each $i^{th}$ finger ${x}^{(t,i)}\in\RR^{D_x/N_\mathsf{fingers}}$ to high dimensional features $x^{(t,i)}_\mathsf{feat}$ via a learnable function $f_\mathsf{finger}:\RR^{D_x/N_\mathsf{fingers}}\mapsto \RR^{H'}$, i.e., $x^{(t,i)}_\mathsf{feat} = f_\mathsf{finger}({x}^{(t,i)}),~\forall i\in[N_\mathsf{fingers}]$. These features are then concatenated with the previous hidden state $h^{(t-1)}$ to produce a hand feature $x^{(t)}_\mathsf{feat}$ via another learnable function $f_\mathsf{hand}:\RR^{N_\mathsf{fingers}H'+H}\mapsto \RR^{H}$. The final feature $x^{(t)}_\mathsf{feat}$ is concatenated with $y^{(t-1)}$ and fed to the LSTM cell to update the hidden state ${h}^{(t)}$ (see \cref{fig:enc_arch}). Due to the usage of LSTM cells, the hidden state we pass between FACs keeps track of all the past input. In this way, all past operations $\by^{(1:t-1)}$ are considered at time $t$ with more focus on recent steps, creating the temporal dependency we desire. Note that in \cref{fig:enc_arch}, the internal structure of finger feature module is omitted for clarity.

Now, with the feature $\bh$ extracted from input data, we can summarize $\bh$ and $\by$ using an aggregation function $a$, and then generate the final Gaussian posterior $q_\phi(z\mid \bx,\by)$ with mean $\mu_\phi$ and covariance $\Sigma_\phi$. This procedure follows \cite{anp} and can be summarized as follows:
\begin{align}
    s_C &= a(f_\mathsf{a}(h^{(1)},y^{(1)}),\dots,f_\mathsf{a}(h^{(N)},y^{(N)})) \label{eq:sC_fa} \\
    \mu_\phi,\Sigma_\phi &= f_\mathsf{\phi}(s_C). \label{eq:f_phi}
\end{align}
Here, $f_\mathsf{a}:\RR^{H+D_y}\mapsto \RR^H$ is a learnable feature extractor providing extra flexibility. $a:\RR^{N\times H}\mapsto\RR^{H}$ can be any function that reduces the time dimension, e.g., mean function. $f_\phi:\RR^{H}\mapsto \RR^{D_z}\times\RR^{D_z}$ is a learnable function that produces the final statistics of latent $z$, assuming $\Sigma_\phi$ is a diagonal matrix.

\subsection{Decoder $p_\theta$}\label{sec:cchp_decoder}

Next, we explain the computation carried out by the decoder, shown in \cref{fig:CCHP_arch} (right). Using the context hidden states $\bh_C$, context handling operations $\by_C$, and latent $z_C$ sampled from the approximate prior $q_{\phi\mid C}$, the decoder's task is to predict the handling operations $\by_T$ for a target commands $\bx_T$. The explicit dependency of $\Pi_\theta$ on user context (see section \ref{sec:cchp_intro}) is encoded in $(\bh_C,\by_C)$ pair, while the temporal dependence in $\by_T$ is captured by a recurrent structure similar to that in the encoder. The basic unit of the decoder is a \textit{decoder cell} which processes one target command $x_T^{(t)}$ at a time to predict $y_T^{(t)}$, while considering previous handling operations $\by_T^{(1:t-1)}$ and user context. Next, we present the structure of decoder cells.

\paragraph*{\textbf{Decoder Cell} $f_\mathsf{dec}$}
 
Shown in \cref{fig:dec_arch}, the decoder cell is first composed of the same encoder cell $f_\mathsf{enc}$ used in the encoder. This encoder cell similarly captures the dependency between handling operation $y_T^{(t)}$ at current timestep and past trajectory $\by_T^{(1:t-1)}$ (see Eq.~\eqref{eq:generative_process}) and maintains a target hidden state $h_T^{(t)}$. Now, we feed $h_T^{(t)}$ into a \textit{context attention} module to generate a hidden representation $r^{(t)}$ that captures useful information for current time step from the context. Context attention, as in \cite{anp}, learns to attend over context hidden states $\bh_C$, known as keys, for each target hidden state $h^{(t)}_T$, known as a query. This is done by mapping both to the same embedding space with a learnable function $f_{\mathsf{kq}}$ and then selecting similar context-target states in dot product sense. Such similarities are treated as weights and used to compute a weighted sum over the context handling operations $\by_C$:
\begin{align}\label{eq:context_att}
    \lambda_{u}^{(t)} &= \mathrm{softmax}(\langle f_{\mathsf{kq}}(h_C^{(u)}), f_{\mathsf{kq}} (h^{(t)}_T)\rangle),~\forall u\in[N_C].
\end{align}
Then, we have $r^{(t)} = \sum_{u=1}^{N_C} \lambda_{u}^{(t)} y^{(u)}_C$. In \cref{fig:dec_arch}, $f_\mathsf{kq}$ is not shown, but it is applied to the hidden states (i.e., on $\mathsf{key}$ and $\mathsf{query}$ edges) before they enter the context attention block. The dot product then determines whether a given context and target point share the same motion aspect. $r^{(t)}$ should thus capture the handling operations in context that are most relevant to the current target timestep. Intuitively, the attention module represents a similarity measure for dynamic gestures and extract insights from user-speicifc data to support online prediction. Finally, we feed the hidden representation $r^{(t)}$, current hidden state $h_T^{(t)}$ and sampled latent $z_C$ into a learnable function $f_y$ to predict the distribution over the handling operation for this timestep, ${y}_T^{(t)}$ as follows
\begin{equation}\label{eq:f_y}
    \mu_\theta^{(t)}, \Sigma_\theta^{(t)} = f_y(r^{(t)}, h_T^{(t)}, z_C).
\end{equation}

To this end, we have described the neural network architecture of CCHP implementation. Note that, CCHP is derived based on novel intuitions in a theoretical sense. Based on that, a carefully designed NN architecture is equally crucial for CCHP to be effective in real-world scenarios. Hence, we present the preceding sections in a great level of details to show how different CCHP features are practically achieved by architecture designs. For example, we achieve adaptation to different users via the context attention module and achieve robustness to noisy input via the recurrent structure in both the encoder and the decoder.  Our architecture is designed such that it can serve as a practical reference for other task settings when similar features are desired. 

\subsection{Training CCHP with Teacher Forcing}

Note that in sections \ref{sec:cchp_highlvl} to \ref{sec:cchp_decoder}, we present CCHP as in testing time. In this section, we also consider training and denote the model-predicted target operation as $\hat{y}_T^{(t)}$ to distinguish from the observed values (ground truth) $y_T^{(t)}$. According to Eq.~\eqref{eq:generative_process}, when predicting $\hat{y}_T^{(t)}$, we should use ground truth $y_T^{(t-1)}$ from the previous step. In the scenario where Eq.~\eqref{eq:generative_process} can be modelled perfectly and the initial handling operation $y_T^{(0)}$ can be assumed to be zero motion, we could use Eq.~\eqref{eq:elbo_2} directly to optimize our model. However, due to noises from collected data and the inductive bias introduced by our model architecture, our neural network cannot serve as a perfect, universal function approximator. As a result, during testing, the first prediction $\hat{y}_T^{(1)}$ will have error, and this error will propagate to the next timestep's prediction as in $\hat{y}_T^{(2)} \sim p_\theta(y_T^{(2)}\mid \hat{\by}_T^{(1)}, x_T^{(2)},\bx_C,\by_C,z)$ where we include $\hat{y}_T^{(1)}$ in the past trajectory $\hat{\by}_T^{(1)}$. Such error propagation will build up for future timesteps, leading to a large covariate shift between the training and online distribution over $\by_T^{(1:t-1)}$. This would ultimately cause the model to fail during testing.

In addition, there are robustness issues \cite{bengio2015teacherforce,chan2015listen} when LSTMs are used (in encoder cells). Specifically, LSTMs strongly rely on the dependency between consecutive time steps. If trained only with ground truth as input, the model will always assume that its previous output is correct and cannot handle errors in previous steps. At test time where only the model's own output is available, minor errors in predictions can compound on one another, causing robustness issues. As a remedy, we apply a practical modification to the training procedure called \textit{teacher forcing}, initially noted by \cite{bengio2015teacherforce} and popularized by \cite{chan2015listen}. This technique allows the model's own prediction $\hat{y}_T^{(t-1)}$ to be fed to next time step with some probability $1 - p_\mathsf{TF}$, shown as the additional green cell in \cref{fig:dec_arch}. The probability $p_\mathsf{TF}$ of using ground truth gradually decreases throughout training. With that, the model initially learns to generate reasonable handling operations in short term, and then work on long-term prediction to gain robustness against its own previous errors. In our model, we reduce $p_\mathsf{TF}$ at a fixed, linear rate.

% through a training technique known as Teacher Forcing, such recurrent networks suffer from robustness issues at test time if only trained using ground truth values as $y_T^{(t-1)}$.

% Feeding in ground truth values encourages model dependency on the previous output.

% \ruic{btw*2, if you look at Eq (9), you will find it possible to sample z not from each steps' posterior, but from the posterior given pure ground truth data. I guess this still maintains the benefits of TF (feeding predicition in the conditions), but gets closer to the theoretical target (sampling z from true posterior in training)?}
% \ashek{Hm I see, sampling z from the full true posterior is what we did before right? Now we sample from the posterior of each step.}

% Dummy LSTM: MLP's for object and hand feature extraction, two-layer LSTM temporal feature extraction and output generation

% \input{content/6_cobot_task.tex}

\section{Experiments and Empirical Results}

In previous sections, we have introduced RTCoHand framework for cobot handling (section \ref{sec:rtcohand}),  proposed CCHP to model the human-robot interaction (section \ref{sec:model_theory}), and provided a neural network-based implementation for the cobot handling policy (section \ref{sec:arch}). In this section, we put the pieces together and show how to learn the cobot handling policy from human-human demonstrations. We proceed by first introducing the data collection in policy customization (section \ref{sec:collect_user_policies}). Then, we present quantitative evaluations (section \ref{sec:quant_eval}) on our model as well as baselines and ablation models using different metrics (described in section \ref{sec:baseline_ablation_eval_metric}). We then interpret our model qualitatively in section \ref{sec:qual_result}. Finally, we provide a robustness analysis (section \ref{sec:robustness}) of our cobot policy against various human input noises.

\subsection{Collection of Human-Human Handling Demonstrations}
\label{sec:collect_user_policies}

In this section, we describe the procedures and physical setup for collecting training and testing data from human-human collaboration (see section \ref{sec:framework_in_user}). For each  \textit{user} whose command policy is of interest, we involve a \textit{human handler} who takes the role of cobot. The data collection is composed of recording several \textit{clips}, each of which contains a dynamic gesture $\bx$ and expected handling operation $\by$. As preparation of each clip, we sample an operation $\by^* \sim \cY^N$ and instruct the user via a simulated animation of applying $\by^*$ to a virtual object (see \cref{fig:rtcohand_info}). With that as the target, the user performs a self-designed dynamic gesture command $\bx$, while the human handler maneuvers an object to label $\by$ accordingly in the meantime. The user can communicate with the handler at any time to ensure correct interpretation of dynamic gestures.

The user dynamic gestures are detected using OpenPose library \cite{simon2017hand, openpose} with dual Intel RealSense cameras\footnote{https://github.com/IntelRealSense/librealsense} and saved as 3D locations of hand skeleton key points. The object poses are recorded by tracking an ArUco marker \cite{aruco1, aruco2} attached to it and saved as 6D Cartesian poses. Importantly, we point out that $\by^*$ is only a hint on the general object translation and rotation directions, and the user does not need to replicate the exact velocity and duration. Hence, we record the human-maneuvered object motions $\by$ as ground truth, as that represents the desired handling operations in actual human-human collaboration. In addition, we emphasize that we collect data in individual clips to avoid burdening users with the need to rehearse long commands. At test time, our model is able to handle continuous user commands of arbitrary durations since no assumptions on command length has been made. We will verify this in later sections (\ref{sec:hw_demo_example}). See \cref{fig:rtcohand_info} for an illustration.

With the aforementioned procedure, we can collect demonstration data from human-human collaboration and learn customized policies. Notably, to obtain such policies over large dynamic gesture space $\cX^N$ and operation space $\cY^N$, large amount of $(\bx,\by)$ need to be sampled. This is impractical and cognitively demanding because there are infinitely many possible object motions. To mitigate that, we incorporate two empirical simplifications: (a) \textit{dominant motion decomposition} which constrains the operation space $\cY^N$ and (b) \textit{reference initial gesture} which provides a good initial point $x^{(1)}$ (see definition in section \ref{sec:rtcohand_terminology}) to constrain the dynamic gesture space $\cX^N$.

\paragraph*{\textbf{Dominant motion decomposition}}

This feature reduces the amount of object motion that users need to focus on. We observe that most user efforts are spent on finding comfortable gestures to represent translations and rotations along arbitrary \textit{directions}. In comparison, the associated \textit{scales} can be changed by merely adjusting the speed and duration of hand motions and do not require additional design efforts from the user. As such, we ask users to only focus on object motions $\by$ with respect to three \textit{dominant axes}: the horizontal axis (``left and right''), the forward axis (``forward and backward''), and the vertical axis (``up and down''). We will refer to the horizontal, forward, and vertical axes as the X, Y, and Z axes respectively in later sections. We refer to the resulting object motions as \textit{dominant motions}, denoted as $\cY_d^M$. We further define the translation or rotation directions that users intend to achieve as \textit{active dimensions}. For example, if a user intends to move the object to the right (X-axis positive) while rotating it clockwise (Y-axis positive), the recorded operation $\by$ should contain higher values in the first and fifth Cartesian dimensions (active dimensions) than the others (inactive dimensions).
In our human study, all $15$ users agree that focusing on only dominant motions is significantly easier than working on arbitrary object motions. Theoretically, such simplification still preserves the flexibility of possible handling operations because rigid body transformations are decomposable. That is, for all $\by\in\cY^N$, we can find $\by_d\in\cY_d^M$ such that applying both operations to an object would land it in the same final pose. In our empirical study, we also find evidence of such decomposition in human policies, i.e., users are able to maneuver the object to any position and orientation they desire by stacking dominant motions. Nevertheless, it is worth doing a comprehensive investigation on how much task flexibility is preserved when using $\by_d\in\cY_d^M$ only. We leave that for future work.

\paragraph*{\textbf{Reference initial gesture}}

Although the preceding feature greatly reduces the complexity of policy customization, we still find it time-consuming to design dynamic gestures for certain object motions due to limited joint flexibility. To further ease this process, we make an empirical suggestion that the users initiate every dynamic gesture with their hands roughly facing downwards and relaxed, as if they were resting their hands on a desk. This gesture can be viewed as a ``neutral'' state that leaves room for various hand joint movements and is likely to lead to comfortable dynamic gestures. Note that it only describes a general mental image rather than any specific gesture for users to replicate. Users can implement the actual gesture in any way they see intuitive and comfortable. For example, when they need room for rotating the hand to the right, they might start with fingers pointing towards left. In our human study, users can find comfortable dynamic gestures in first few attempts if starting with the reference initial gesture. Users also agree that receiving less guidance would make the process considerably more challenging, while following more specific guidance would make them feel constrained. As such, we conclude that the reference initial gesture is effective in the sense that it points to a versatile initial point in $\cX$ that facilitates further hand motions.

\begin{figure}[h]
    \centering
    \includegraphics[width=\linewidth]{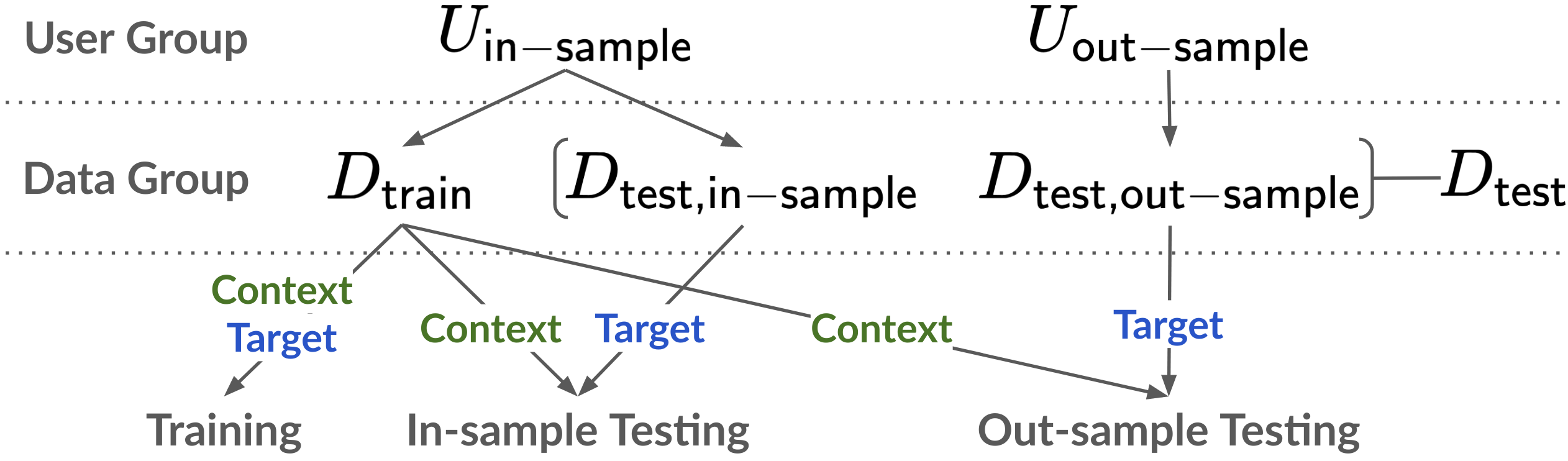}
    \caption{Relations between user, data, and train test settings. Demonstrations in each data group is customized by users from an assigned user group, as indicated by the arrows on top. During training and testing, the context and target data are taken from different data groups, as indicated by the arrows at the bottom.}
    \label{fig:user_data_relation}
\end{figure}

Using the aforementioned procedures, we collect human-human collaboration with $15$ users\footnote{The data collection process only involves two humans performing hand motions and maneuvering a light-weight paper box. The harm or discomfort anticipated are no greater than those ordinarily encountered in daily life.} in total, $10$ assigned to $U_\mathsf{in-sample}$ and $5$ assigned to $U_\mathsf{out-sample}$. From each user, we collect $72$ data clips, each containing a dominant motion that lasts for at most $5$ seconds. For $\mathsf{out-sample}$ users, all $72$ clips are directly assigned to the test set $D_\mathsf{test, out-sample}$. For $\mathsf{in-sample}$ users, data clips are split into a training set $D_\mathsf{train}$ and testing set $D_\mathsf{test, in-sample}$ (see Appendix \ref{append:train_test_data} for details). $D_\mathsf{test, in-sample}$ serves as a validation set used to fine-tune our hyperparameters. $D_\mathsf{test,out-sample}$ allows us to evaluate model generalization across a new group of users $U_\mathsf{out-sample}$ whose unique command styles were never observed at training time. Overall, $D_\mathsf{test} \defeq D_\mathsf{test, out-sample} \bigcup D_\mathsf{test, in-sample}$. See \cref{fig:user_data_relation} for an illustration of the relations between user groups and data groups. 
% This is summarized by the following list:

% \begin{enumerate}
%     \item $D_\mathsf{train}=\{(\bx_\mathsf{train}, \by_\mathsf{train}) : (\bx_\mathsf{train}, \by_\mathsf{train}) \sim u_i; u_i \in U_\mathsf{in-sample} \}$.
%     \item $D_\mathsf{test, in-sample}=\{(\bx_\mathsf{test}, \by_\mathsf{test}) : (\bx_\mathsf{test}, \by_\mathsf{test}) \sim u_i; u_i \in U_\mathsf{in-sample}\}$
%     \item $D_\mathsf{test, out-sample}=\{(\bx_\mathsf{test}, \by_\mathsf{test}) : (\bx_\mathsf{test}, \by_\mathsf{test}) \sim u_i; u_i \in U_\mathsf{out-sample}\}$ 
% \end{enumerate}

% This means that test performance using $\mathsf{in-sample}$ data may be higher than performance on $\mathsf{out-sample}$ data since the model has already trained on the $\mathsf{in-sample}$ user command styles. 
We use $D_\mathsf{train}$ to train our model according to Eq.~\eqref{optim:cobot_goal_2_train}. During testing with $U_\mathsf{in-sample}$ user data, we sample targets $(\bx_T, \by_T)$ from $D_\mathsf{test, in-sample}$ and context data $(\bx_C, \by_C)$ from $D_\mathsf{train}$ and finally evaluate according to Eq.~\eqref{optim:cobot_goal_2}. During testing with $U_\mathsf{out-sample}$ user data, we draw target samples from $D_\mathsf{test, out-sample}$ and context from $D_\mathsf{train}$ and evaluate again using Eq.~\eqref{optim:cobot_goal_2}. See \cref{fig:user_data_relation} again for an illustration. For more details on data collection and training, we refer readers to Appendix \ref{append:train_test_data} and \ref{append:model_training} respectively.

\begin{figure*}[ht]

% \begin{subfigure}[b]{0.09\linewidth}
% \centering
% \includegraphics[width=\linewidth]{img/overall_loss.png}
% \caption{\ashek{Use the special tickz to plot? Also fix legend }  Test loss versus training step.}
% \label{fig:loss}
% \end{subfigure}
% \hfill
% \begin{subfigure}[b]{0.99\linewidth}
\centering

\resizebox{\textwidth}{!}{
\begin{tabular}{lcccccc}
    \toprule
    Models & [D+,U+] & 
                 [D+,U-] & 
                 [D-,U+] & 
                 [D-,U-] &
                 Unseen & 
                 Noisy\\
    \midrule
    Dummy LSTM & (NA, 4.89) & NA & NA & NA & (-360.35, 5.48) & (NA, 5.38)\\
    
    RANP  & (-410.66, 1.66) & (-401.42, 1.95) & (-400.76, 1.96) & (-400.71, 1.96) & (-369.26, 2.54) & (\textbf{-288.59}, 4.84)\\
    
    CCHP (1.0, 1.0) & (-239.82, 1.8) & (-13.61, 2.29) & (577.09, 2.95) & (618.49, 3.04) & (465.95, 3.01) & (121.30, 2.52)  \\
    
    CCHP (0.75, 0.75) & (-611.87, \textbf{1.36}) & (-538.34, 1.62) & (-498.34, 1.68) & (-495.1, 1.7) & (\textbf{-370.33}, 2.31) & (-222.07, \textbf{2.25})  \\
    
    CCHP (0.5, 0.5) & (\textbf{-634.45}, 1.4) & \textbf{(-566.96, 1.62)} & \textbf{(-558.65, 1.62)} & \textbf{(-556.71, 1.63)} & (-346.86, \textbf{2.26}) & (-273.71, 2.28) \\
    
    CCHP (0.1, 0.1)  & (-601.4, 1.43) & (-538.96, 1.66) & (-537.42, 1.66) & 
    (-537.22, 1.66) & (-350.8, \textbf{2.26}) & (-191.37, 2.41)\\
    
    CCHP $p_\mathsf{TF}=0.1$ & (-270.68, 1.96) & (-58.68, 2.28) & (-51.03, 2.32) & (-50.26, 2.32) & (248.07, 2.9) & (805.37, 3.24)\\
    
    CCHP $p_\mathsf{TF}=0.5$ & (-323.25, 1.71) & (-172.0, 1.99) & (-135.91, 2.01) & (-143.76, 2.01) & (50.87, 2.64) & (721.69, 3.58)\\
    
    CCHP $p_\mathsf{TF}=0.9$ & (9512.59, 8.92) & (10554.34, 9.37) & (11120.19, 9.37) & (11050.91, 9.4) & (10000.92, 9.14) & (10354.68, 9.15) \\
    \bottomrule
\end{tabular}
}
% \caption{Final test loss across different ablation studies.}
% \label{table:loss}
% \end{subfigure}
\captionof{table}{Test losses of different models (rows) across different test settings (columns). ``CCHP'' refers to our main model with $(p_{\mathsf{D+}},p_{\mathsf{U+}}) = (0.5,0.5)$ and $p_\mathsf{TF} = 0.9$ with linear decay. ``CCHP (*,*)'' refers to ablation model with $(p_{\mathsf{D+}},p_{\mathsf{U+}}) = (*,*)$ and $p_\mathsf{TF} = 0.9$ with linear decay. ``CCHP $p_\mathsf{TF}=*$'' refers to ablation models with $(p_{\mathsf{D+}},p_{\mathsf{U+}}) = (0.5,0.5)$ and a fixed $p_\mathsf{TF} = *$ rate. All results are shown as (ELBO loss, MSE loss), the lower the better. The lowest losses under each test setting are marked in bold. D+/D- means the active dimension in target is/is not contained in the context. U+/U- means the target comes from the same/different person as/than the context.}
\label{table:loss}
\end{figure*}

\subsection{Baselines, Ablations, and Evaluation Metrics}\label{sec:baseline_ablation_eval_metric}
To examine the importance of different components of our model, namely context attention and the encouragement of temporal dependency in the ouput, we compare our model's performance with two baseline models respectively:
\begin{enumerate}
    \item \textit{Can we interpret dynamic gestures without any context?} We compare to Dummy LSTM, an MLP-based hand feature extractor followed by a two-layer LSTM. Dummy LSTM notably does not use any context during prediction and optimize the negative log likelihood instead of ELBO due to lack of latent posterior.
    \item \textit{Can we solve the task without temporal dependency in the output?} We compare to RANP \cite{qin_recurrent_2019}, a standard ANP \cite{anp} extended with recurrent feature extraction: a sliding window LSTM. RANP notably does not encourage temporal dependencies in the output. RANP optimizes the same ELBO loss as ours (see \eqref{eq:elbo_3}).
\end{enumerate}

We also analyze the impact of key hyperparameters. First, we can decide what context to be provided for each target sample at training. For a given target sample $(\bx_T, \by_T)$, we generate the context samples $(\bx_C, \by_C)$ based on two criteria: the source user and the active motion dimensions. Context can be taken from the same (U+) or different user (U-) as the target with some probability $p_{\mathsf{U+}}\in(0,1)$. The context can also contain the same (D+) or different active dimensions (D-) as those of the target, and this is chosen with probability $p_{\mathsf{D+}}\in(0,1)$. Intuitively, context from the same user can contain useful information about the user's unique hand motion styles, even if its active motion dimensions are different. When training our model, for each target clip, we provide one context clip chosen by the above procedure (according to $p_{\mathsf{D+}}$ and $p_{\mathsf{U+}}$) and two other context clips drawn randomly.

In most scenarios, all timesteps of the sampled context clips are used. However, if a context clip is both taken from the same user and contain the same active dimensions (essentially being the exact same data as the target), we follow \cite{anp} to only preserve a slice of the target $(\bx_T, \by_T)$ in context. To achieve that, we randomly sample a starting timestep and duration: 
\begin{align}
    (\bx_C, \by_C) \leftarrow (\bx_T, \by_T)[t_0:t_0+K] \nonumber \\
    \text{ subject to $t_0\geq 0;\, K\geq T_\mathsf{min};\, t_0+K  \le p_\mathsf{cxt}|\bx_T|$}
\end{align}
where $T_\mathsf{min}$ is a minimum length hyperparameter, $|\bx_T|$ is the length of the target sample $\bx_T$, and $p_\mathsf{cxt}$ specifies the maximal proportion of the target that can be made context.
% In addition to this target-based context, we append $2$ full-length context clips that do not contain any of the target sample's active dimensions. This ensures that context is never empty. 

Our main CCHP model is trained with an initial teacher forcing rate of $p_\mathsf{TF} = 0.9$, which is constant for the first 600 training steps and linearly decreases thereafter. We set constant values of $p_{\mathsf{D+}} = 0.5$ and $p_{\mathsf{U+}} = 0.5$. Ideally, the model is provided context that is relevant to interpreting the given target sample. However, since we desire online inference, we have a computational constraint defined by the length of context that may prevent the context from containing any relevant data. Hence, during training, we only provide relevant context from time to time based on probabilties ($p_{\mathsf{D+}} < 1$ and $p_{\mathsf{U+}} < 1$). To examine the impact of these key hyperparameters, we compare performance with several ablations that share the same architecture as CCHP, but only differ from our main model by one hyperparameter:
\begin{enumerate}
    \item\textit{Is a curriculum for teacher forcing necessary to train a robust model?} We train three CCHP models with fixed Teacher Forcing rates: CCHP $p_\mathsf{TF}=0.1$, CCHP $p_\mathsf{TF}=0.5$, and CCHP $p_\mathsf{TF}=0.9$ while keeping all other parameters unchanged.
    \item \textit{Does the choice of context also impact our model's reliance on context?} We train a CCHP model with different context sampling probabilities $(p_{\mathsf{D+}}, p_{\mathsf{U+}})$: $(0.1, 0.1)$, $(0.75, 0.75)$, and $(1.0, 1.0)$.
\end{enumerate}

At test time, we evaluate the performance of baselines and our model on two separate testing sets $D_\mathsf{test, in-sample}$ and $D_\mathsf{test, out-sample}$. Similar to training, both testing sets are composed of target pairs $(\bx_T, \by_T)$. However, unlike during training where we configure the context using some probabilities $(p_{\mathsf{D+}}, p_{\mathsf{U+}})$ both in $(0,1)$, we consider the extremes of picking context by setting either or both probabilities to $0.0$ or $1.0$. This enables us to explicitly examine the model's reliance on different context.
\begin{enumerate}
    \item \textbf{Same user same dimension (D+,U+)}: The target active dimensions are \textit{always} contained in context taken from the \textit{same} testing user. 
    \item \textbf{Same user different dimension (D-,U+)}: The target active dimensions are \textit{never} contained in context taken from the \textit{same} testing user.
    \item \textbf{Different user same dimension (D+,U-)}: The target active dimensions are \textit{always} contained in context taken from a \textit{different} testing user.
    \item \textbf{Different user different dimension (D-,U-)}: The target active dimensions are \textit{never} contained in context taken from a \textit{different} testing user.
    \item \textbf{Unseen user (unseen)}: The target motions are taken from $D_\mathsf{test, out-sample}$, which was collected from users not involved in training. Hence, the context always comes from different users than the target.
\end{enumerate}

Finally, we examine whether our CCHP model is robust against noisy human input as motivated in section \ref{sec:cobot_goal}. To do that, we simply perturb the target dynamic gestures $\bx_T$ during testing with Gaussian noises. We will provide a more extensive study of the robustness in section \ref{sec:robustness}.
\begin{enumerate}
    \item[6)] \textbf{Noisy input}: Gaussian noise is added to the input hand motion with zero mean and standard deviations of $\sigma_T = 0.050$ and $\sigma_R = 0.025$ for translation and rotation respectively. The context is not perturbed and is chosen from the (U+, D+) setting. 
\end{enumerate}

\begin{figure*}[ht]
    \centering
    \begin{subfigure}[b]{0.56\linewidth}
        \includegraphics[width=\linewidth]{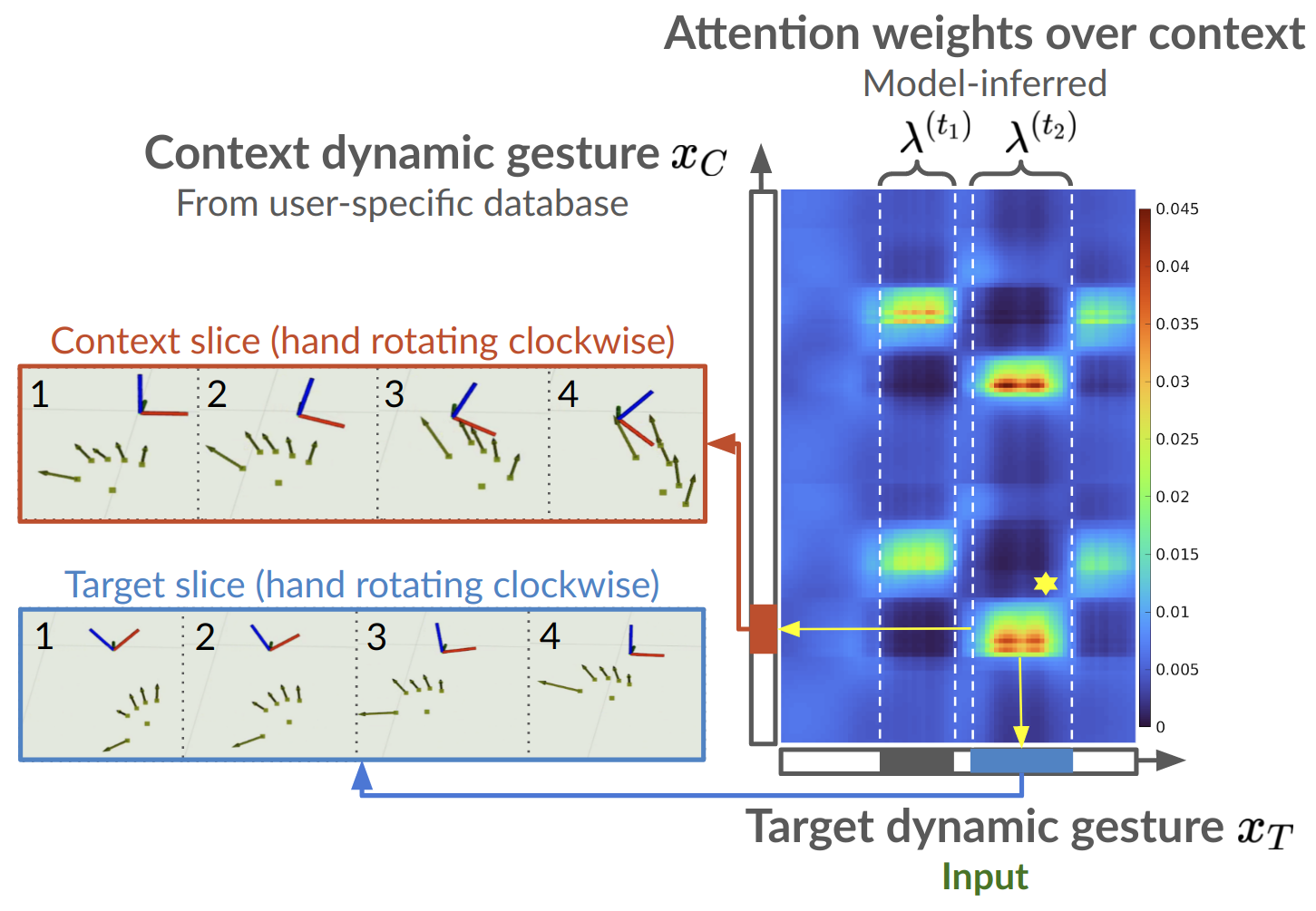}
        \caption{Context-target attention.}
        \label{fig:qual_cxt_att_left}
    \end{subfigure}%
    \hfill
    \begin{subfigure}[b]{0.44\linewidth}
        \includegraphics[width=\linewidth]{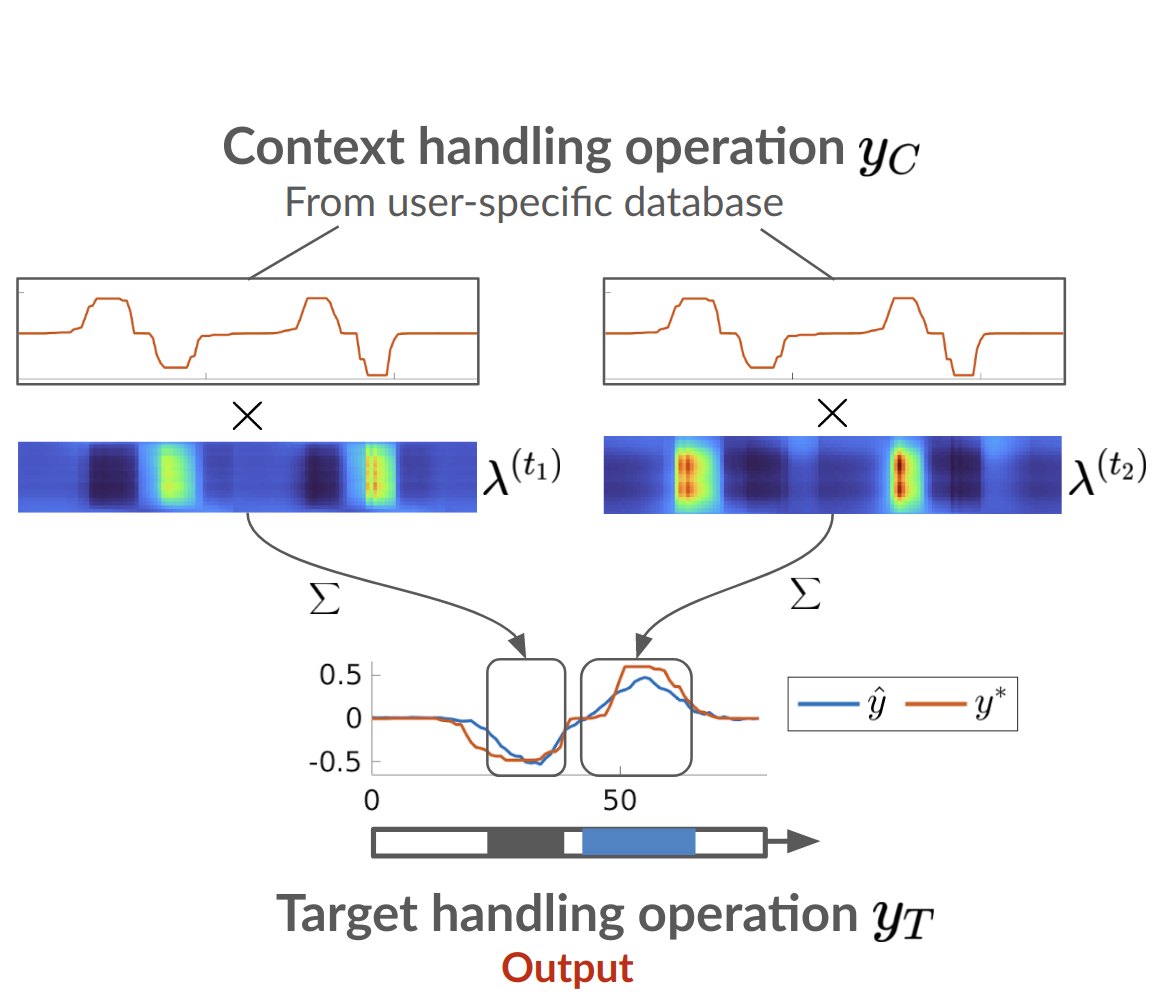}
        \caption{Computation diagram for target output.}
        \label{fig:qual_cxt_att_right}
    \end{subfigure}%
    \caption{Computation diagram of CCHP with data visualization. Sub-figure (a) visualizes the attention weights (color map) of target dynamic gestures $\bx_T$ over context $\bx_C$. Sub-figure (b) shows the computation of target object motion $\by_T$ based on context $\by_C$. In (a), the context time axis is vertical pointing upwards. The target time axis is horizontal pointing to the right. Two slices of dynamic gestures are visualized, one each from context and target. The hand is plotted as arrows representing the principle direction of fingers and a point for the palm. The axis represents the object pose with X, Y, Z axes in red, green, and blue respectively. The Y-axis is pointing forward in compliance with the right-hand rule. Sub-figure (b) conceptually shows the computation of two target operation slices from attention weights ($\lambda^{(t_1)}$ and $\lambda^{(t_2)}$) and context operations (top). The ground truth target operation is also shown (red). All operations' plots only show the rotation velocities in Y-axis. Note that the regions on the horizontal axes marked in black and blue refer to the same time ranges in target dynamic gestures (input) and handling operations (output) in (a) and (b) respectively. }
    \label{fig:qualitative_cases}
\end{figure*}

\subsection{Empirical Quantitative Evaluation}\label{sec:quant_eval}
We now evaluate the performance of our baselines on our collected human motion dataset. Table \ref{table:loss} compares model losses on different test datasets averaged across 5 different trained seeds of each model. ``NA'' is placed where the experimental result does not apply. Specifically, the Dummy LSTM does not use context data at all, thus is not affected by the variation of $(p_\mathsf{D+}$, $p_\mathsf{U+})$ settings. Dummy LSTM also does not model latent posterior distribution so the ELBO loss does not apply. 
Looking at each column, our proposed CCHP model achieves the lowest loss ELBO across almost all test settings, showing that our modules contribute meaningfully. The only exception is the Noisy test, where RANP has slightly lower ELBO loss, but much higher MSE loss. We investigate this later in section \ref{sec:robustness}.

As expected, CCHP with a fixed, high teacher force rate of $p_\mathsf{TF}=0.9$ performs poorly at test time since it was mainly given ground truth values for the $y_T^{(t-1)}$. CCHP with a low teacher force rate performs better, but still worse than our main model. This is due to the lack of a curriculum to encourage temporal dependency. CCHP performance does not change much when the $(p_\mathsf{D+}$, $p_\mathsf{U+})$ parameters are varied. RANP performs the closest to our main CCHP model, but still performed worse due to the lack of strong temporal dependency between adjacent predictions $y_T^{(t)}$ and $y_T^{(t-1)}$ as described in Eq.~\eqref{eq:generative_process}.
Comparing performance where context may (D+) or may not (D-) accurately capture the target motion dimensions, we can see that (D+) loss is consistently lower. This shows that providing relevant context data does improve performance. At the same time, performance does not degrade significantly if the context contains irrelevant data. Comparing (U+) to (U-) columns, we also observe that model performance does not worsen if context data is taken from a different user altogether. This also indicates two possibilities: either (a) the $15$ subjects do not differ signficantly in their gesture styles or (b) providing the same dominant motion dimensions is more important than providing data from the same user.
Performance on unseen data is worse than the other settings, which matches our expectations since these users' hand motion policies were never observed during training. However, performance is not much worse for CCHP variants, indicating that our model can successfully generalize to new users. 
Lastly, comparing the columns (D+,U+) and ``Noisy'', we notice worse performance across all models in either ELBO or MSE loss. Robustness to noise relies on two crucial aspects: hidden states temporally consistent between adjacent timesteps and context attention to reference similar hand motion data when the raw input itself may be too noisy to interpret correctly. 
% Looking at the relative percent increase in error:

Comparing ``RANP'' to ``CCHP $p_\mathsf{TF}=0.9$'', both models make predictions approximately independently of previous predictions. ``RANP'' computes all predictions $\hat{y}_T^{(t)}$ as independent outputs of an MLP. ``CCHP $p_\mathsf{TF}=0.9$'' has a teacher force rate of nearly $p_\mathsf{TF}=1.0$, meaning that it was trained to make predictions $\hat{y}_T^{(t)}$ conditioned on ground truth $y_T^{(t-1)}$ and not on its own prediction $\hat{y}_T^{(t-1)}$. However, ``CCHP $p_\mathsf{TF}=0.9$'' performs significantly worse because it assumes the previous output is accurate and is not robust to its own prediction errors.

% \ashek{More quantitative analysis once we have finalized loss table}
% \ruic{Robustness against noise is related to pTF, so temporal structure is critical; same person same motion context helps; adaptation case are similar; unseen is worse}

\subsection{Empirical Qualitative Results}\label{sec:qual_result}

% \subsubsection{Case \#1}

% We first provide an example of model context attention in Fig. \ref{fig:case_1_context_attention} and output in Fig. \ref{fig:case_1_finger_attention} where rotation about the y-axis (Ry) was the dominant motion. In this sample, there are three context motion clips in total. Only the first two clips (starting from the top) contain active rotation about the y-axis. 

We now analyze our CCHP model qualitatively by decomposing the forward computation and visualizing key components such as input dynamic gestures, context attention weights, and output handling operations (see \cref{fig:qualitative_cases}). We aim to show the internal process of CCHP model and interpret how it adapts to the user by inferring meaningful context attention weights.

For clarity, we choose a test case where the ground truth target operation has Y-axis rotation as its dominant motion. We provide two context clips which also contain Y-axis rotations. Recall that the CCHP model first generates a context hidden state $\bh_C$ using the encoder (section \ref{sec:cchp_encoder}) from the context data $(\bx_C,\by_C)$. Then, the model processes the input, target dynamic gestures $\bx_T$, into target hidden state $\bh_T$ (section \ref{sec:cchp_decoder}). Then, using Eq.~\eqref{eq:context_att}, the model computes context attention weights $\blambda\defeq \{\lambda^{(t)}_u\}_{t\in[N_T],u\in[N_C]}$ for each target time step $t$ over each context time step $u$. The resulting attention weights $\blambda$ is shown in \cref{fig:qual_cxt_att_left}. For a given target timestep (on horizontal axis), the model places normalized attentions across the context timesteps (vertical axis). Now, we focus on one region of high attention marked by a yellow star. This region shows high attention at the target time range marked in blue over the context time range marked in red. Visualizing the corresponding dynamic gestures, we see that they are indeed similar: in both cases, the hand is rotating clockwise. In those frames, we also show the object poses which composes Y-axis rotations. Note that the object poses in the target slice is the ground truth and is not available to the model.

With context attention $\blambda$, the model proceeds to generate target handling operation $y_T^{(t)}$ at each time step $t$. To illustrate, we focus on two target time ranges marked in black and blue in \cref{fig:qual_cxt_att_left} horizontal axis and their attentions $\lambda^{(t_1)}$ and $\lambda^{(t_2)}$ over the context (marked by white dashed lines). We take out both weight slices and align them with the context handling operation $\by_C$ as shown in \cref{fig:qual_cxt_att_right}. The model calculates a weighted sum of $\by_C$ using $\lambda^{(t_1)}$ and $\lambda^{(t_2)}$ and generate target operation $\by_T$ in corresponding time ranges, as bounded by black rectangles on the plot and marked in black and blue on the target time axis. We can see that in both ranges, the target output is highly related to the context at locations with high attention weights. This indicates that our model is able to locate relevant context portions for prediction at different time steps. Note that in this test case, we demonstrate a single-dimensional target operation. Our model is also able to compose complex motions with multiple active dominant dimensions by placing attentions on various context clips. This will be illustrated further in section \ref{sec:hw_demo_example} with a real-time cobot handling task.

\begin{figure}
    \centering
    \resizebox{0.9\linewidth}{!}{\input{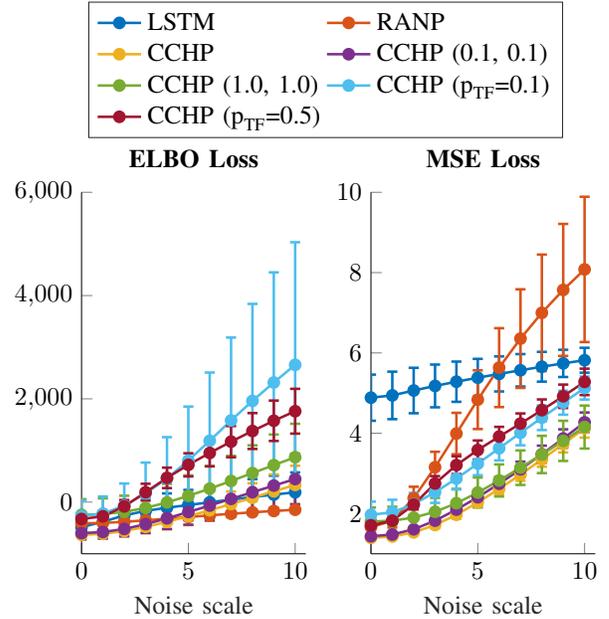}}
    \caption{ELBO and MSE testing loss (D+,U+) under $11$ levels of input noises (starting from noise-free). Mean and one standard deviation over five repeated evaluations are shown for each noise level.}
    \label{fig:loss_vs_noise}
    \vspace{-10pt}
\end{figure}

% \subsubsection{Case \#2}

% We now consider the setting where context is irrelevant to the target and does not contain any active motion dimensions of the target. Looking at the left context plot in Fig. \ref{fig:case_2_context_attention}, the context's active dimension was only rotation about the x-axis (Rx), which explains why the plot of Tx velocity is nearly zero for all timesteps. We can also see this visually in the four context frames, where the hand mainly rotates about the red x-axis. Visually, this does not match the motion of the target frames, which mainly contain translation along the red x-axis and rotation about the blue z-axis. This helps explain why the context attention looks fairly uniform for translation along x: no context timesteps are particularly relevant. Nonetheless, the model still generates the correct magnitude of translation along the target timesteps. 
\begin{figure*}
    \centering
    \resizebox{0.9\linewidth}{!}{\input{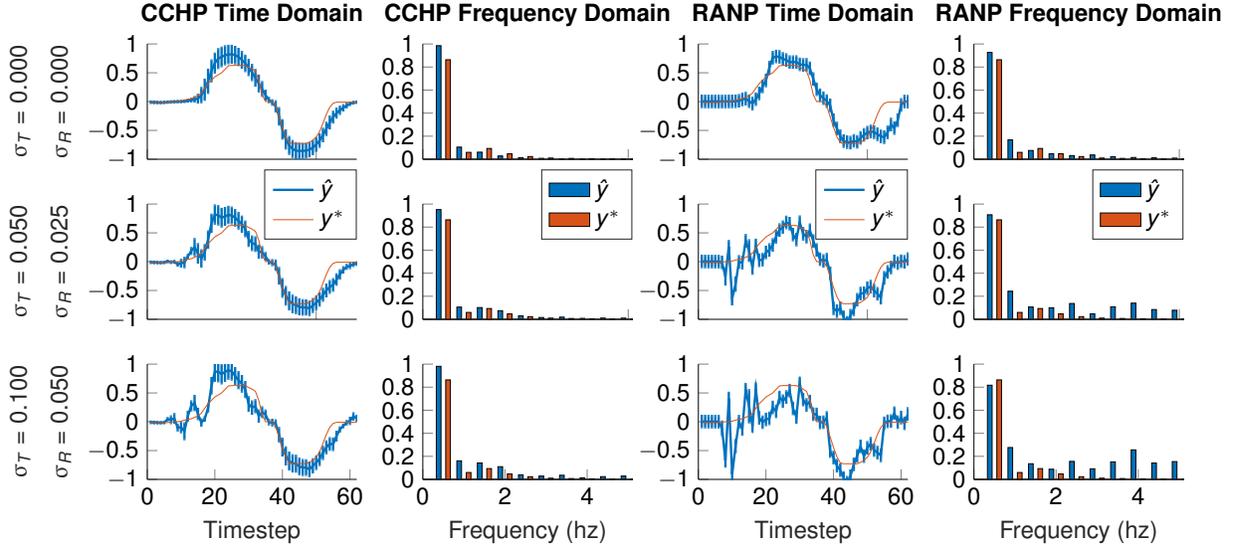}}
    \caption{Prediction of CCHP and RANP under different levels of input noises in both time and frequency domains. Both model prediction and ground truth on $X$ axis rotation are plotted. Model predictions are plotted in mean and one standard deviation. Each row corresponds to a level of input noise, given by the standard deviation of Gaussian noises (marked to the left) added to translation and rotation velocities of target dynamic gestures. }
    \label{fig:freq_analysis}
\end{figure*}

\subsection{Robustness Against Noisy Human Motions}\label{sec:robustness}

In section \ref{sec:quant_eval}, we have shown the testing performance of all baseline and ablation models under a fixed level of artificial input noise. In this section, we conduct a more extensive study with various levels of input noises. Specifically, we perturb the human dynamic gestures using zero-mean Gaussian noises with standard deviation $\sigma_T$ for translation velocities and $\sigma_R$ for rotation velocities. We use $10$ levels of noises that increases from $(\sigma_R=0.0,\sigma_T=0.0)$ to $(\sigma_R=0.1,\sigma_T=0.05)$ with equal increments. We evaluate all baseline and ablation models using the same user same motion (D+,U+) test setting and summarize their performances in \cref{fig:loss_vs_noise}. It is evident that in both ELBO and MSE loss evaluations, our main model shows the top robustness against increasing noise intensities. Interestingly, for some models, their ELBO and MSE losses do not follow the same trend. For example, the ELBO loss of RANP model appears to grow the slowest with larger noises, but its MSE loss grows the quickest. Note that although the ELBO loss indicates the model's fitness of data, it is not as directly related to the real-world cobot performance as MSE loss. This is because the MSE loss examines the mean of cobot policy prediction, which represents the most confident action that the cobot would take during actual deployment.

To help illustrating the discrepancy between two evaluation metrics, we visualize the predictions of both RANP and our model on the same test case under three input noise levels: $(\sigma_R=0.0,\sigma_T=0.0)$, $(\sigma_R=0.05,\sigma_T=0.025)$ and $(\sigma_R=0.1,\sigma_T=0.05)$. These settings correspond to none, medium, and top noise levels in \cref{fig:loss_vs_noise}. We also plot the model predictions and ground truth in frequency domain to see whether the input noises would cause unwanted high frequency components in the output. We can see from the time domain plot that RANP predictions becomes significantly noisier than CCHP (ours) results as we increase the input noise level. Such prediction noises are reflected only in MSE loss but not ELBO loss, leading to the discrepancy between those two metrics. Similarly in frequency domain plots, we can see that the input noise leads to non-trivial high frequency components of RANP predictions that are not present in the ground truth. While in CCHP predictions, the frequency components are more consistent with the ground truth with minor unwanted high frequency components. Thus, we conclude that our model can robustly generate smooth handling operations under non-trivial input noises, which can be ubiquitous in real world scenarios due to human motion uncertainty and perception noises.

\section{Real-Time Cobot Handling Task}\label{sec:rt_cobot_handling_task}

In pervious sections, we have shown the performance of our cobot policy based on human-human collaboration data. Now, we apply the cobot policy to a real-time human-robot collaboration task, \textit{collaborative inspection}, to demonstrate our work in real world scenarios. In this task, a human worker needs to insepct a newly molded metal workpiece from different view angles. Due to high temperature, the human will command a cobot to maneuver the workpiece during inspection. In the following sections, we first ellaborate on the inspection task and corresponding hardware setup (section \ref{sec:hw_demo_task}). To demonstrate the accessibility and flexibility of RTCoHand framework, we conduct a baseline study where the human user commands the robot with a joy stick controller instead of dynamic gestures. Then, we compare our approach to the baseline on both objective measures (e.g., completion time) and subjective metrics (e.g., how easy to use) (section \ref{sec:hw_demo_exp}). We also show a sample of real-time robot actions and context attentions in section \ref{sec:hw_demo_example}.

\subsection{Collaborative Inspection and Hardware Implementation}
\label{sec:hw_demo_task}

\begin{figure}[h]
    \centering
    \begin{subfigure}[b]{.49\linewidth}
        \centering
        % {\setlength{\fboxsep}{0pt}
        \includegraphics[width=\linewidth]{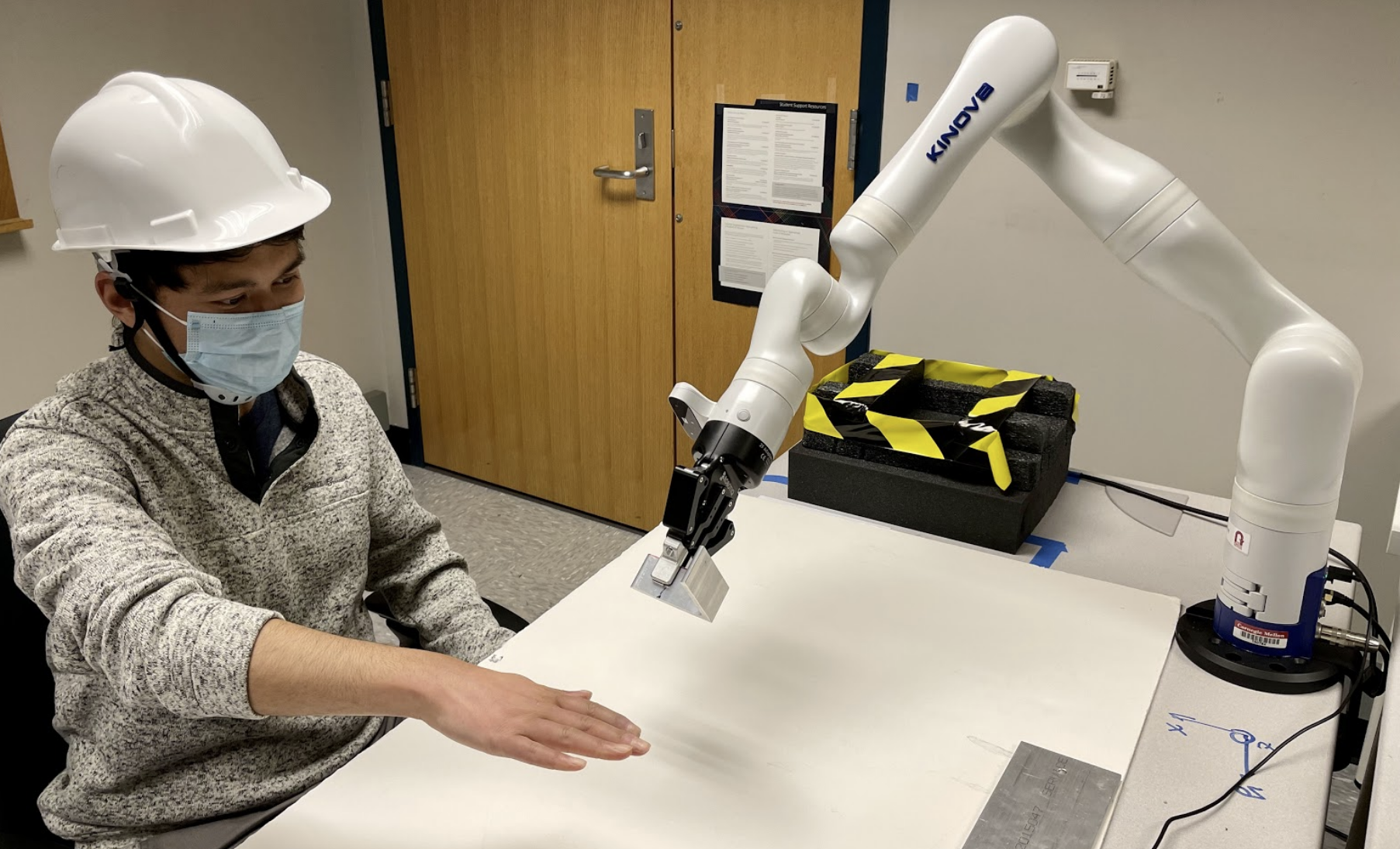}
        \caption{RTCoHand}
        \label{fig:metal_inspect_hand}
    \end{subfigure}%
    \hfill
    \begin{subfigure}[b]{.49\linewidth}
        \centering
        % {\setlength{\fboxsep}{0pt}
        \includegraphics[width=\linewidth]{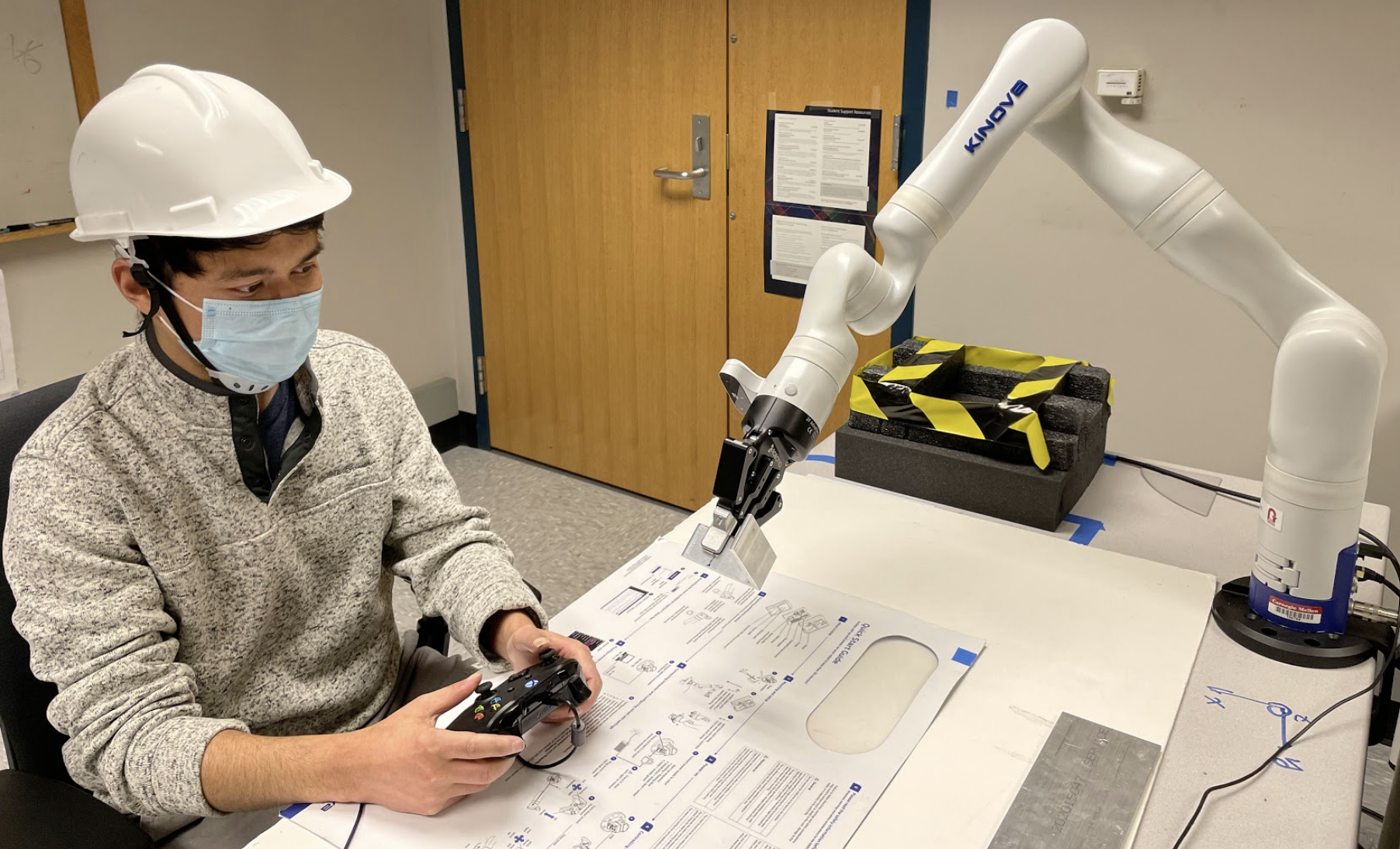}
        \caption{Joy-stick controller}
        \label{fig:metal_inspect_controller}
    \end{subfigure}%
    \vfill
    \begin{subfigure}[b]{.49\linewidth}
        \centering
        % {\setlength{\fboxsep}{0pt}
        \includegraphics[width=\linewidth]{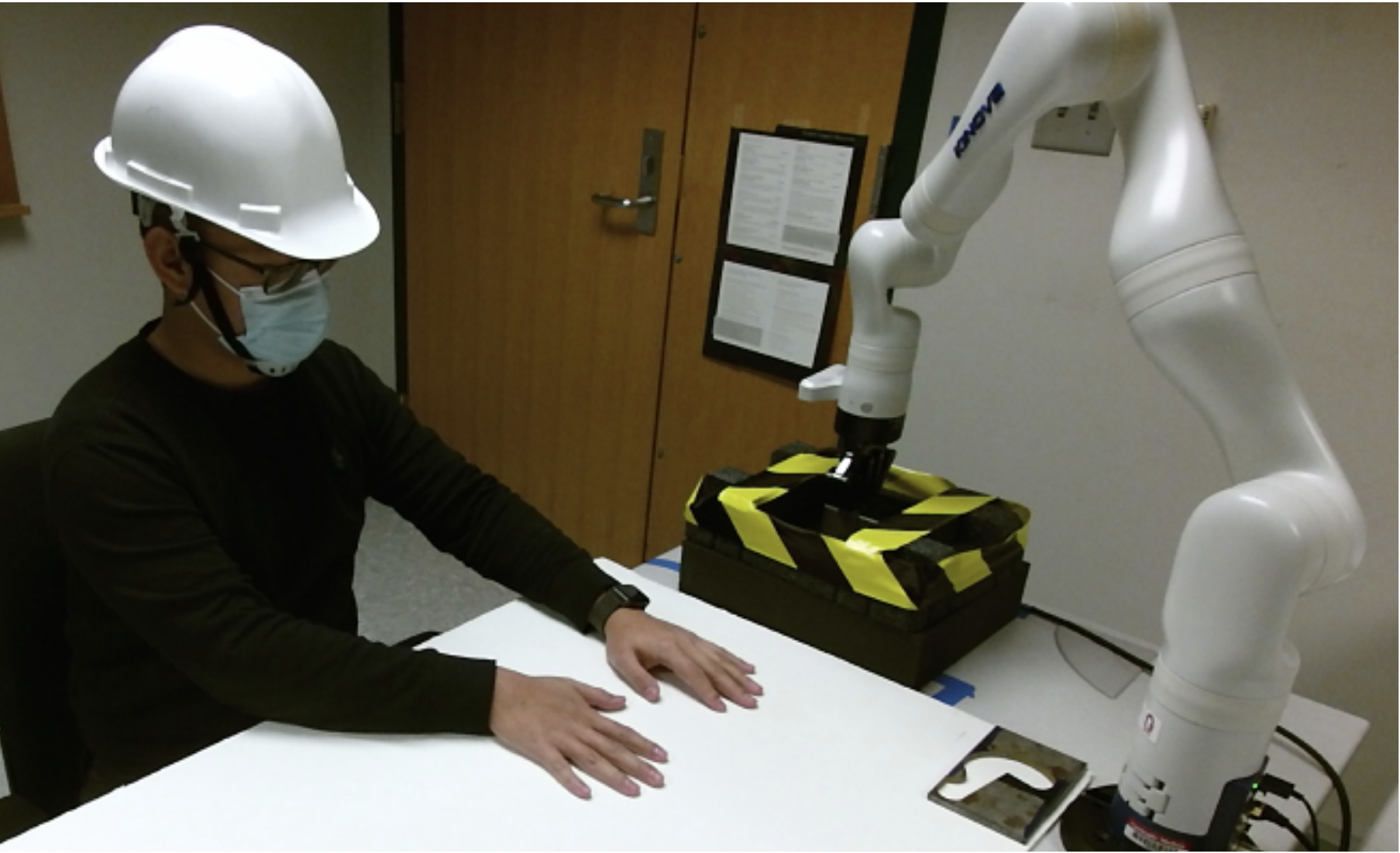}
        \caption{Initial task state}
        \label{fig:metal_inspect_init}
    \end{subfigure}%
    \hfill
    \begin{subfigure}[b]{.49\linewidth}
        \centering
        % {\setlength{\fboxsep}{0pt}
        \includegraphics[width=\linewidth]{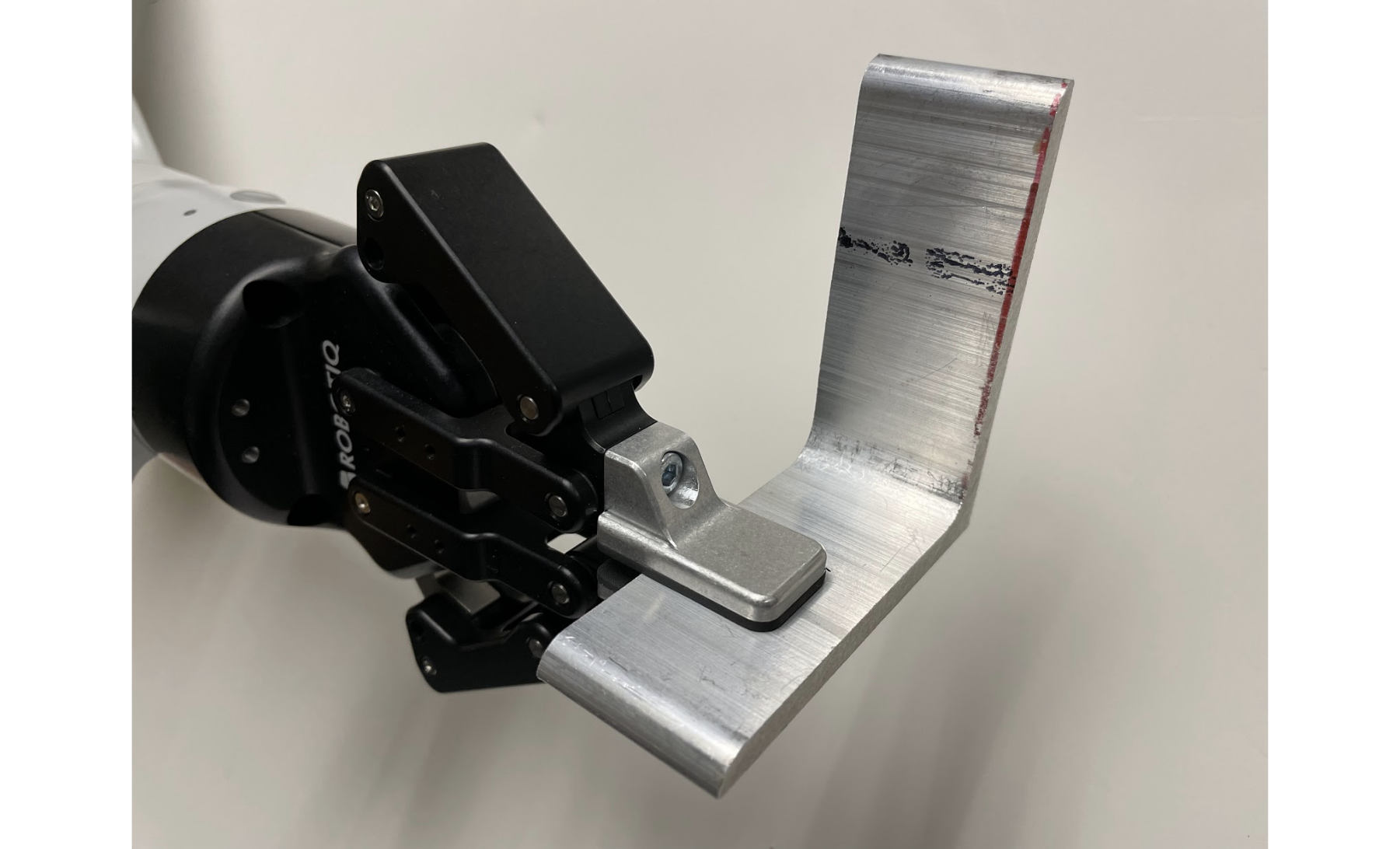}
        \caption{Metal workpiece}
        \label{fig:metal_inspect_component}
    \end{subfigure}
    \caption{Photos of collaborative inspection task. The human users can interact with the cobot using either (a) RTCoHand framework or (b) joy-stick controller. Initially, the metal workpiece (d) is placed in a mold marked with black and yellow tape as shown in (c). Users need to first lift the workpiece out of the mold and perform inspection.}
    \label{fig:hw_demo_task}
\end{figure}

In our inspection task, a human worker needs to inspect all surfaces of a newly molded metal workpiece (see \cref{fig:hw_demo_task}). At the beginning, the newly molded workpiece rests in the mold and is already grapsed by the cobot (see \cref{fig:metal_inspect_init}). Due to high temperature, the worker cannot directly handle the workpiece (\cref{fig:metal_inspect_component}). Instead, the worker controls a cobot to first lift the workpiece out of the mold and maneuver it for inspection (\cref{fig:metal_inspect_hand} and \cref{fig:metal_inspect_controller}). In this work, we demonstrate a ``safe version'' of the task using a small piece of room-temperatured metal and a fake mold (taped in black and yellow in \cref{fig:metal_inspect_init}). Notably, this ``safe version'' is still highly representative of real manufacturing scenarios since the same setting easily applies to workpieces of arbitrary size, shape, and properties. In particular, when the newly molded workpiece is large, it might only be safe for workers to inspect from behind a safety glass shield instead of walking around. To reflect such constraint in our study, we ask human subjects to perform the inspection with minimal body movement. For better view angles, they would rely on the cobot handler to move and rotate the workpiece. We apply two approaches to this task: dynamic gesture-based control with RTCoHand framework and joy stick controller-based control from robot arm manufacturer. Next, we describe their hardware implementation.

% At each time step, we  $q_{\phi\mid C}$
% \begin{align}\label{eq:generative_process}
%     & p_\theta(\by_T\mid \bx_T,\bx_C,\by_C)\nonumber \\
%     \defeq \int & p_\theta(\by_T\mid \bx_T,\bx_C,\by_C,z) p(z) dz. \nonumber \\
%     = \int & \prod_{t=1}^{N_T} p_\theta(y_T^{(t)} \mid \by_T^{(1:t-1)}, x_T^{(t)},\bx_C,\by_C,z) p(z) dz.
% \end{align}

\paragraph*{\textbf{RTCoHand (ours)}}
To implement RTCoHand framework, we use the same perception system as discussed in section \ref{sec:collect_user_policies}) to capture dynamic gestures, a desktop to handle all computations (hand detection, cobot policy $\Pi_\theta$, etc.), and a Kinova Gen3 robot arm to execute target handling operations (see \cref{fig:metal_inspect_hand}). During the task, dynamic gestures $\bx_T$ are detected and sent to the desktop at $10$ Hz. This frequency is limited by our computational resources, and can be higher with enhanced hardware. Upon receiving $\bx_T$, the desktop invokes the learned cobot policy $\Pi_\theta$ and generate target handling operations $\by_T$ in the same frequency $10$ Hz, where the latent variable $z$ is sampled from $q_{\phi\mid C}$ (see the end of section \ref{sec:learn_infer_CCHP}). To promote safety, we send the operations to the robot for execution at a lower frequency $2$ Hz, with each command being the average of the past $10$ predicted handling operations and upper bounded by conservative thresholds. This post-processing stage is equivalent to first applying a moving average filter to robot actions, then performing a down sampling. Note that this post-processing is not necessary to our RTCoHand framework, and is incorporated only as a practical safety measure. In scenarios where workers and cobots are apart, the safety measure can be relaxed or lifted to make the cobot more responsive. During the task, we also ensure that the metal workpiece is always securely held by a gripper installed on the cobot, and that the human user is able to pause and resume cobot actions when desired (e.g., to re-position the hand for new command). Every time the cobot resumes, we re-sample the latent variable $z$ from $q_{\phi\mid C}$ and use that sample until the next pause. In this work, for simplicity, the user can pause the cobot by verbally instructing a human operator who has access to the desktop. In practice, the human operator can be steadily replaced with a voice recognition-based switch that is integrated in the system.

% \subsection{Safety of human users}

% Although RTCoHand framework allows contact-free handling of objects, it is likely that the robot is still operating in proximity with human users. There are several approaches available for ensuring human safety from cobot's side, such as using safe planning to generate collision-free trajectories or monitoring contact via force sensors. In RTCoHand framework, we want to add safety measures accessible from users' side. Specifically, we allow users to disable the cobot with a manual switch when they do not want the cobot to interpret their motions as handling commands. For example, users might be working on the workpiece without needing to move it, or might need to reset their hands to avoid reaching joint limits in new dynamic gestures. In practice, the manual switch can be implemented via a physical device easy to access during interaction, such as a foot pedal or a portable button.

\paragraph*{\textbf{Controller (baseline)}}

To implement a baseline approach, we swap the RTCoHand framework with a joy stick controller provided by the robot arm manufacturer (see \cref{fig:metal_inspect_controller}). With the controller, the worker can move and rotate the robot end-effector at a factory-defined speed. In case of Kinova Gen3 robot arm, the controller does not support moving and rotating simultaneously, so the worker needs to toggle between position and orientation control modes. The worker also has access to the manufacturer-provided operation manual during the task.

\begin{figure*}[ht]
    \centering
    \begin{subfigure}[l]{0.9\linewidth}
        \includegraphics[width=\linewidth]{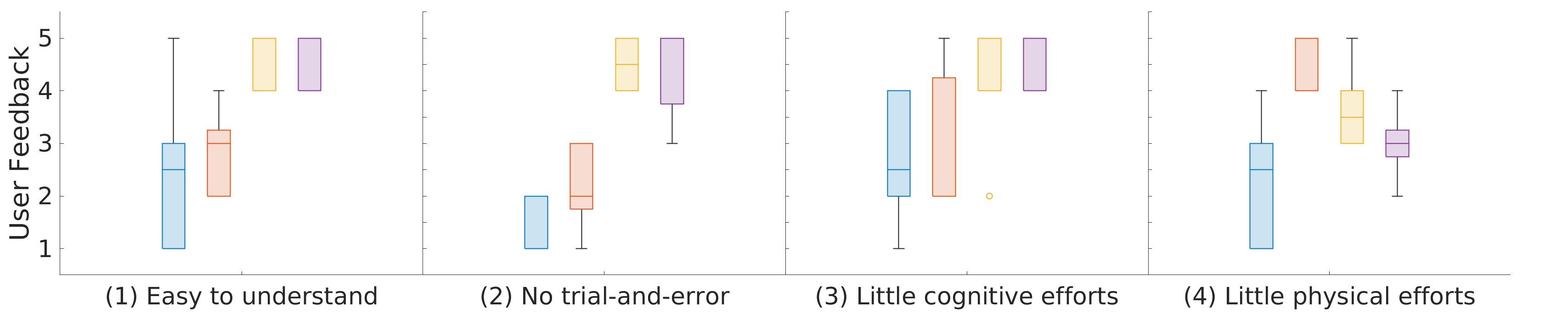}
    \end{subfigure}%
    \hfill
    \begin{subfigure}[l]{0.9\linewidth}
        \includegraphics[width=0.92\linewidth]{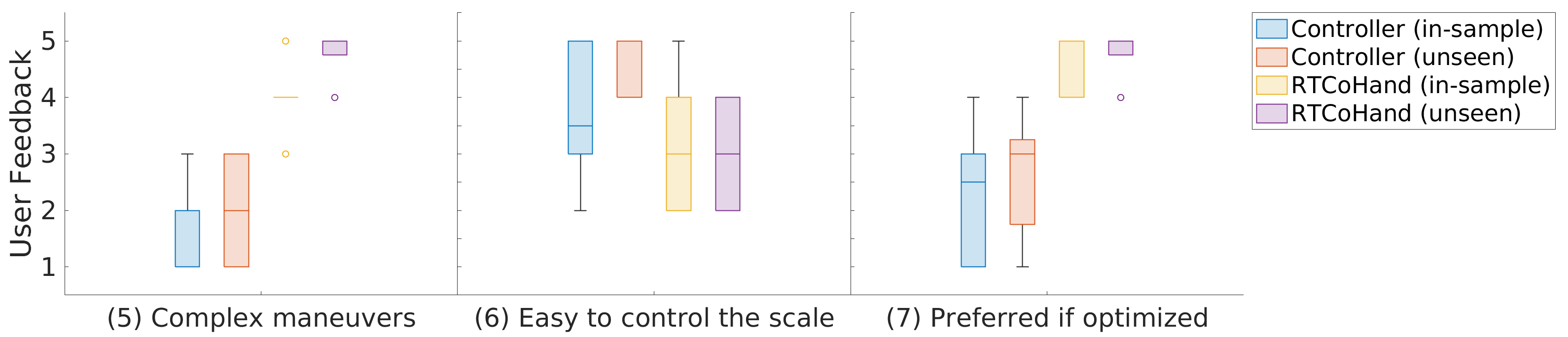}
    \end{subfigure}%

    \caption{User evaluation on collaborative inspection task on seven metrics as introduced in section \ref{sec:hw_demo_exp}. Metric (1) to (4) correspond to the accessibility category. Metric (5) and (6) correspond to flexibility category. Metric (7) corresponds to feasibility category. All evaluations are on five-point scale where $1$ means ``disagree'' and $5$ means ``agree''. We group the feedbacks in both interaction approach and user types, resulting in four groups in total.}
    \label{fig:hw_user_feedback}
\end{figure*}

\subsection{Experiments and Evaluations}
\label{sec:hw_demo_exp}

We study how RTCoHand framework performs with two types of human users\footnote{In this experiment, we deploy safety measures including bounded robot speed and safety zones that keep the robot away from human subjects by a minimum distance. All invited subjects are students within the institute who fully understand the stated safety measures before consenting to the experiment. The harm or discomfort anticipated are no greater than those ordinarily encountered in daily life.}: \textit{in-sample} and \textit{unseen}. For in-sample group, we invite $6$ users from $U_\mathsf{in-sample}$ in the human-human collaboration process (see section \ref{sec:collect_user_policies}). For unseen group, we invite $5$ users new to the cobot handling task. This simulates the real world situation where new workers need to directly work with existing robotic systems. In-sample users are familiar with the cobot handling task, but have not interacted with real cobot. In addition, user-specific context data is only available for in-sample users while context for unseen users are randomly chosen from in-sample user database. All users also repeat the inspection task with joy-stick controller.

With each approach, each human subject is first given $3$ minutes to explore and get familiarized with the control interface as well as cobot actions. Then, all users perform the same collaborative inspection task. To access the task quality, we report time of completion as an objective metric and user feedback on different task aspects as subjective metrics. Specifically, we ask users to evaluate both RTCoHand and controller on seven metrics in three categories: accessibility, flexibility, and overall feasibility. Answers are given on a five-point scale where $1$ means ``disagree'' and $5$ means ``agree''. See following for the questions and their intuitions. Users are also free to provide extra descriptive feedbacks on any other aspects not already covered.

\paragraph*{\textbf{Accessibility}}
Recall that an HRC task is accessible if it requires little tehnical skills to understand and easy to perform. We evaluate this via the following four metrics.
\begin{enumerate}
    \item \textit{It is easy to understand how to control the cobot.}
    \item \textit{I do not need trial-and-error during interaction.} Here, trial-and-error means trying different inputs until the cobot move and/or rotates in the intended direction(s).
    \item \textit{It takes little cognitive efforts to control the cobot.}
    \item \textit{It takes little physical efforts to control the cobot.}
\end{enumerate}

\paragraph*{\textbf{Flexibility}}
An HRC task is flexible if human users can control the cobot to achieve a wide range of operations. We evaluate this via the following two metrics.
\begin{enumerate}
    \item[5)] \textit{I can control the cobot to do complex maneuvers (e.g., move and rotate the object at the same time towards arbitrary directions).}
    \item[6)] \textit{When the cobot moves or rotates the workpiece in my intended direction, I can easily control the amount it moves.}
\end{enumerate}

\paragraph*{\textbf{Feasibility}}
In this category, we ask about the general feasibility of the cobot helper in real-world production. Namely, we ask the users to evaluate in terms of solely the interaction approach, assuming the hardware implementation is optimized to industrial standard. We evaluate this via the following question.
\begin{enumerate}
    \item[7)] \textit{If the cobot system is highly optimized (e.g., highly responsive and does not lag behind the control input), I prefer \textbf{this} method in everyday production tasks.} ``\textbf{This}'' refers to either RTCoHand or controller. Note that there can be many criteria in ``industrial standard''. For simplicity, we use the responsiveness of the cobot as a proxy since the lag between user input and cobot actions is the most frequently mentioned issue in extra user feedbacks.
\end{enumerate}

See \cref{fig:hw_user_feedback} for a visualization of the user evaluation scores. Comparing the score distributions of different groups on each metric, we can gain useful insights on how RTCoHand can be beneficial in real-world scenarios, how it works with experienced and new users, and how we can potentially improve. We summarize our findings as follows.

\begin{enumerate}
    \item RTCoHand provides a significantly easier and more intuitive way to interact with cobots, as evident in metric (1) of \cref{fig:hw_user_feedback}.
    \item Examining metric (2), we find that using the controller, most users need to find the correct input operation though trial-and-error. This is mainly due to the non-intuitive mapping from input operations such as ``push the left stick'' to desired object motions such as ``rotate clockwise''. With RTCoHand, users rarely need to do that due to the highly intuitive way of control-by-hand and customized dynamic gestures. We emphasize that in most real-world scenarios, trial-and-error is unacceptable, since wrong commands can be expensive or even fatal (e.g., when handling large components such as ship parts or hazardous meterials such as corrosive liquid). Hence, we anticipate RTCoHand to be beneficial when control errors are to be avoided.
    \item As shown in metric (3), as expected, users agree that RTCoHand is not mentally demanding to use. They have mixed opinions on the controller, depending on whether they have prior experience with controllers or not. However, it is consensus that RTCoHand requires less cognitive efforts to work with.
    \item From metric (4), we see that all users have mixed opinions on both approaches with respect to required physical efforts. While using the controller does not need obvious body movements, some users need to move their head back and forth to examine the metal workpiece and find the correct stick or button on the controller. While commanding by dynamic gestures require body movemenst, some users learn to save efforts by resting their elbows on the table as support. Hence, we leave the study of systematically saving physical efforts for future work.
    \item Metric (5) indicates that RTCoHand is more flexible than the controller in supported operations. RTCoHand supports arbitrary object motions, while with the controller, users can only switch between pure translation and pure rotation.
    \item According to metric (6), the controller out-performs RTCoHand in terms of operation precision. This is expected since the controller have an explicit stick or button for each cobot action (e.g., push the left stick to rotate along Y-axis) and users can adjust the cobot's speed by applying appropriate forces on the controller. However, we argue that with minor enhancement, RTCoHand can also achieve equivalent precision. For example, we can allow users to adjust the overall sensitivity of the cobot via a control knob and scale the cobot operations.
    \item Metric (7) indicates the overall user impression on both interaction approaches in real-world scenarios. Users agree that with optimized hardware implementation, RTCoHand would be prefered over classical controllers in everyday production.
\end{enumerate}

\begin{figure*}[ht]
    \centering
    \centering
    \begin{subfigure}[b]{.38\linewidth}
        \centering
        % {\setlength{\fboxsep}{0pt}
        \includegraphics[width=\linewidth]{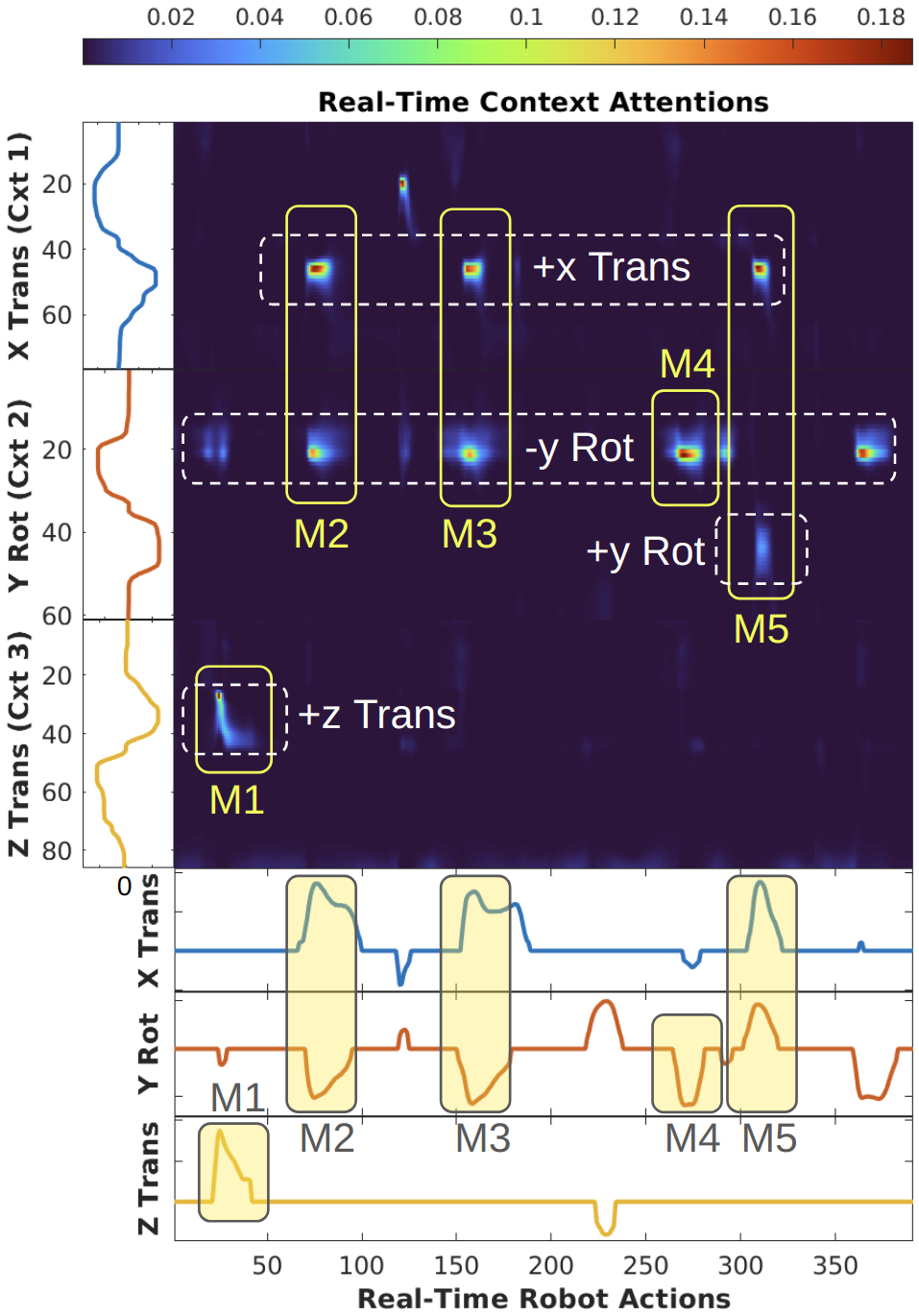}
        \caption{Context, attentions, and robot operations.}
        \label{fig:hw_demo_sample_cxt}
    \end{subfigure}%
    \hfill
    \centering
    \begin{subfigure}[b]{.59\linewidth}
        \centering
        % {\setlength{\fboxsep}{0pt}
        \includegraphics[width=\linewidth]{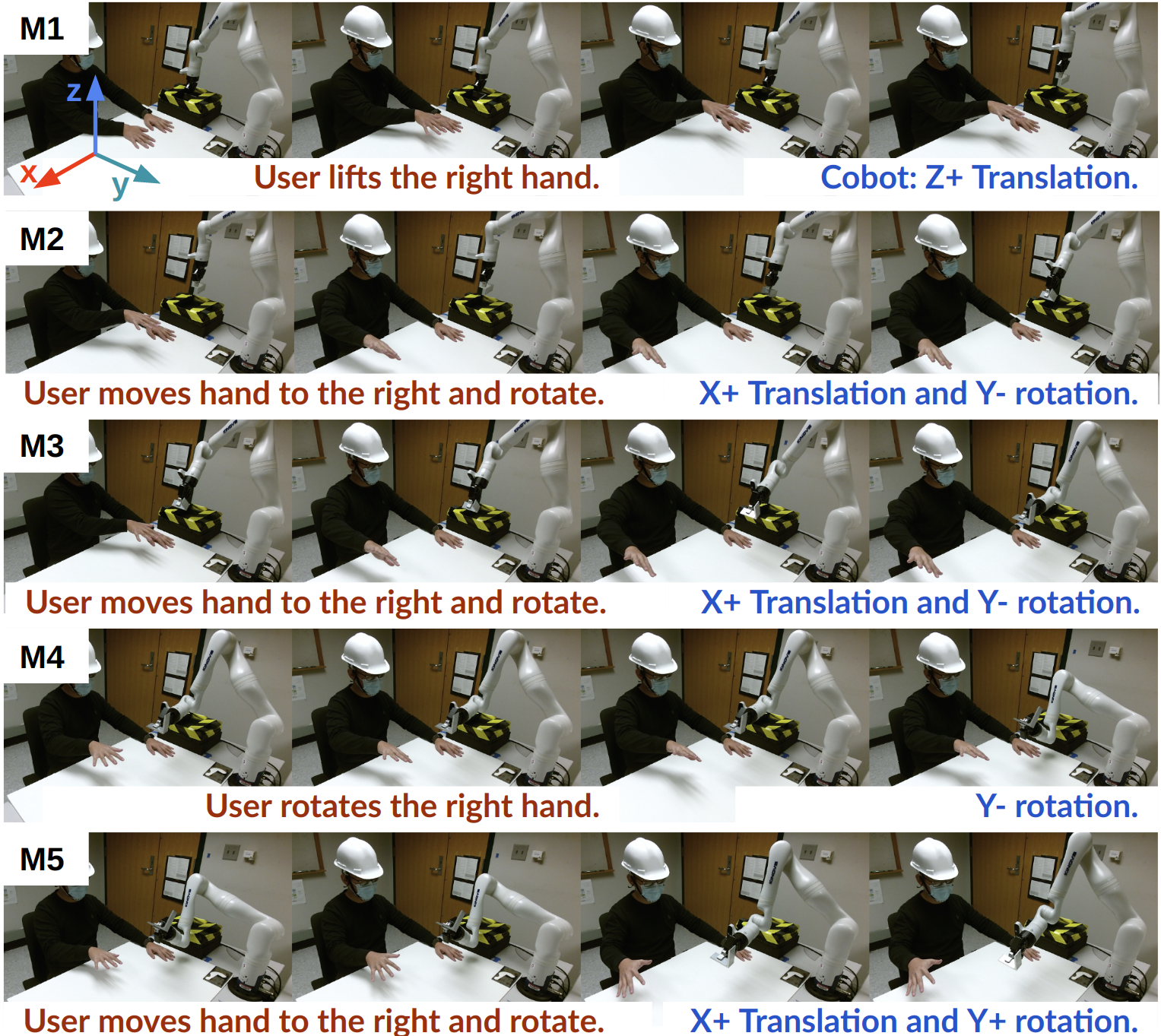}
        \caption{Photos of five handling operations.}
        \label{fig:hw_demo_sample_photo}
    \end{subfigure}%
    \caption{Sample data from real-time collaborative inspection with RTCoHand framework. (a) shows a sample of real-time robot handling operations (bottom) and partial context object motions (left) in three dimensions: X translation, Y rotation, and Z translation. Time axis for context progresses vertically downward while that for real-time actions horizontally to the right. Importantly, the context motions are from three \textbf{different} context clips (Cxt 1, Cxt 2, Cxt 3), each having its own time axis. The attention map (middle) shows the attention weights at each real-time step over each context time step. Active attention locations corresponding to the same context motion are marked with white dashed rectangle. We also mark five cobot handling operations (M1 to M5) composed with one or two active dimensions in yellow rectangle, with corresponding human dynamic gestures shown in (b) (one row per operation). In (b), we also provide a short description of the human dynamic gestures (red) as well as the resulting cobot operations (blue) below each row of photos. The coordinate system setup is overlaid on top left.}
    \label{fig:hw_demo_data}
\end{figure*}

Now, we compare the completion time of collaborative insepction using both RTCoHand and the controller. Recall that the goal is to inspect all surfaces of a metal workpiece. Since we do not specify the order of surfaces to inspect or the time to spend on examining each surface, the completion times among users vary significantly, with minimum and maximum being $50$ seconds and $215$ seconds respectively. Hence, we report the mean and standard deviation of the completion time with RTCoHand normalized by that with the controller for each user: $1.89\pm 0.66$. We identify two major issues contributing to this result. First, our hardware implementation is bottlenecked by the camera systems for detecting hand keypoints. The system currently runs at around $10$ Hz and publishes hand keypoints with a delay of $257.7$ milliseconds on average per frame. This creates lag between user commands and cobot actions and reduces efficiency. Second, the user needs to communicate with another human operator to pause or resume the cobot, which can also be inefficient. These two issues are also frequently mentioned in extra user feedback. However, they deviate from the focus of this work and we aim to resolve both in future work (e.g., with additional computation resources and a voice recognition-based cobot switch).

\subsection{Real-Time Data Sample}
\label{sec:hw_demo_example}

In this section, we interpret the behavior of cobot policy based on a sample of real-time collaborative inspection data, including the cobot actions, context being used, context attention weights, and snapshots of user dynamic gesture commands. In this inspection task, the human worker belongs to out-sample group, meaning that he has participated in the data collection process as discussed in section \ref{sec:collect_user_policies}, but his data is not avaiable during training. Such setting requires the cobot policy to adapt to a new user by corresponding real-time user commands to the given user data that is not involved in training. In following discussions, we shall see that this is indeed the case.

See \cref{fig:hw_demo_data} for visualizations of the data sample. In this task, we provide the cobot policy with six context clips, each of which contains a single active dominant motion (either translation or rotation in either X, Y, or Z axes). Due to space limit, we only focus on the first few hundred time steps of the task where the worker first lifts the workpiece out of the mold and commands a few handling operations. In such a period, the cobot operations mainly involve translations in X anx Z axes and rotation in Y axis. Hence, we only plot real-time cobot actions in those three active dimensions, as shown in \cref{fig:hw_demo_sample_cxt} at the bottom. It immediately follows that only the three context clips containing those active dimensions are of interest, as shown in \cref{fig:hw_demo_sample_cxt} to the left. Note that same dimension in context and real-time actions are plotted in matching colors. Finally, we plot the attention weights at each real-time step over each plotted context frame with the color scales shown on top. Different from \cref{fig:qualitative_cases}, we omit the dynamic gesture axes and directly align the context and target operation plots with attention map. However, note that the attention map is still computed from context and target dynamic gestures, and the real-time (target) robot actions are generated using the attention map. In the next paragraphs, we provide a deep dive to interpret how our cobot policy corresponds real-time user commands to given context, compute reasonable attentions, and generate desired operations.

First, focusing on the attention map, we observe several regions with high attention weights. Regions in the same horizontal line correspond to the same context time step. We mark four such groups using white dashed rectangles and annotate their semantic meanings. For example, the ``-y Rot'' group corresponds to around step $20$ in context clip $2$, where the plot shows negative rotation velocities in Y-axis. On the other hand, regions in the same vertical line correspond to the same target time range. We mark five such groups (in yellow), each considered as an individual handling operation. We also mark those operations in the cobot action plot at bottom and provide real-world photos at those time steps. Each of the marked operations has attention across the context in different locations, indicating different characteristics of real-time dynamic gestures. Also, all operations have different durations, since this is fully decided by the user in real time.

Now, we proceed by focusing on one specific time range during the task: around target time step $320$ where the last operation (M5) takes place. The model places attention mainly around two context locations as marked by yellow rectangle in \cref{fig:hw_demo_sample_cxt}. This indicates that the model identifies two different patterns in the human dynamic gestures $\bx_T$, each matching different context. Observing the real-world photos of operation M5 (\cref{fig:hw_demo_sample_photo} bottom), we can easily identify the combination of two such patterns: \textbf{rotating the hand around Y-axis} while \textbf{moving along X-axis} at the same time. With that in hand, we can verify the correctness of attention weights with the high-attention context steps of M5: around step $50$ in context clip $1$ and step $45$ in context clip $2$. High attention weights mean that the model considers dynamic gestures $\bx_T$ in M5 to be similar to $\bx_C$ within \textbf{both} of those context time ranges. By construction of the context attention, the corresponding context handling operations $\by_C$ should also be reasonable for $\bx_T$ in M5. To verify, we see that the context handling operation ($\by_C$) plot indeed shows an \textbf{X-axis translation} and a \textbf{Y-axis rotation} at those high-attention context steps respectively. Hence, we have verified that our CCHP model can extract \textit{multiple} patterns from the \textit{same} dynamic gesture command $\bx_T$, and correctly correspond each of these patterns to a \textit{different} relevant context time range. Alternatively, one can visualize and compare $\bx_T$ with $\bx_C$ during those context time ranges as what is done in \cref{fig:qualitative_cases}. This would lead to the same conclusion and thus is omitted.

Now, with reasonable attention weights, the composite motion in M5 is finally mapped to the real-time robot action $\by_T$ marked in \cref{fig:hw_demo_sample_cxt} (bottom). As expected, $\by_T$ contains both positive X-axis translation and Y-axis rotation. The same analysis directly applies to all other operations M1 to M4. Thus, we conclude that our cobot policy is able to combine different dominant motions from the context data and generate complex handling operations, when the exact same dynamic gestures in real time are not directly present in the context.

% \paragraph*{\textbf{Case \#2 to add}}

\section{Conclusion and Future Directions}

In this work, we have presented a novel real-time collaborative robot handling (RTCoHand) framework that allows different users to generate complex object handling trajectories using dynamic hand gestures only. We enable the cobot to be both easy to use and flexible in supported operations by allowing users to customize their own control strategies with comfortable dynamic gestures. We have identified several key challenges in implementing RTCoHand framework, including adaptation to different users, human motion uncertainty, and safe and robust robot actions under noisy input. To tackle these challenges, we take a probabilistic view of the cobot handling task and propose conditional collaborative handling process (CCHP) to learn a stochastic cobot policy that conditions on user-specific database. We learn the policy using human-human collaboration data and present both quantitative and qualitative evaluations to verify that all key challenges are resolved. We finally verify our approach on a real-time collaborative inspection task and show that the learned cobot policy is advantageous in both accessibility and flexibility, especially in terms of complex operations, low user cognitive burden, and no need for trial-and-error.

Finally, we point out several directions for future work.
\begin{enumerate}
    \item As discussed in section \ref{sec:rtcohand_terminology}, both operation and command spaces should be constrained in practical human-robot collaboration tasks. In this paper, we apply \textit{ad hoc} constraints that theoretically preserves all task flexibility (see section \ref{sec:collect_user_policies}), which is supported by our real time experiments. However, it requires more extensive studies to show how exactly those constraints affect the accessibility and flexibility of cobot handling tasks. It is worth studying their relations such that different operation and command spaces can be designed according to application-specific requirements. 
    \item In this paper, we apply our method to an inspection task where all operations on the workpiece are done by the cobot. This only represents the most simplistic setting among a wide range of scenarios our method is applicable to. In general, any task that requires object manipulation can benefit from this work, e.g., collaborative assembly where the human user installs components on a large base held by the cobot, and tele-controlled sorting where the cobot manipulates (hazardous) materials and arrange them in a container. As future work, we aim to study the efficacy of our approach on advanced settings with more complex maneuvers and procedures.
    \item As shown in the user study, our hardware implementation is still improvable in various aspects. For instance, the delay between user commands and cobot actions has been frequently mentioned in the user feedback. Besides, the sensitivity of cobot actions to human dynamic gestures is only comfortable to partial users and an adjustable sensitivity level is desired. Lastly, users should be able to pause and resume the cobot conveniently, e.g., via a voice recognition-based switch. As future work, we aim to optimize the implementation of RTCoHand framework on physical cobots.
    % \item Use RNP for modeling changing styles on top of changing actions.
    % \item In this paper, we consider the human users to maintain a fixed command policy. However, this might not hold in human-robot collaboration, since human users may adapt their policies when robot actions are not satisfactory. Hence, it is worth studying how human adapt their command policies, and how cobot policy should take that into consideration.
    % \item Improved hardware and more complicated tasks
\end{enumerate}


%%%%%%%%%%%%%%%%%%%%%%%%%%%%%%%%%%%%%%%%%%%%%%%%%%%%%%%%%%%%%%%%%%%%%%%%%%%%%%%%

%%%%%%%%%%%%%%%%%%%%%%%%%%%%%%%%%%%%%%%%%%%%%%%%%%%%%%%%%%%%%%%%%%%%%%%%%%%%%%%%

%%%%%%%%%%%%%%%%%%%%%%%%%%%%%%%%%%%%%%%%%%%%%%%%%%%%%%%%%%%%%%%%%%%%%%%%%%%%%%%%


\appendix

\subsection{Minimization of KL divergence}\label{append:KL_ELBO}

% We show how one derives \eqref{eq:elbo_1} and \eqref{eq:ELBO_def} from \eqref{eq:KL} as follows.
We now show that minimizing the KL divergence \eqref{eq:KL} is equivalent to maximizing the ELBO as in \eqref{eq:elbo_1} and \eqref{eq:ELBO_def}. Recall the learning objective of CCHP:
\begin{equation}\label{eq:KL_append}
    \minimizewrt{\phi} D_{\mathsf{KL}}(q_\phi(z\mid \bx_T,\by_T,\bx_C,\by_C) \mid\mid p(z\mid \bx_T,\by_T,\bx_C,\by_C))
\end{equation}
First, we have
\begin{align}
    p(z\mid \bx_T,\by_T,\bx_C,\by_C) = \frac{p(\by_T\mid \bx_T, \bx_C, \by_C, z)p(z)}{p(\by_T\mid \bx_T, \bx_C,\by_C)}.
\end{align}
Plugging in the objective of \eqref{eq:KL_append} and expand, we have
\begin{align}
    D_\mathsf{KL} & = \EE_{z\sim q_\phi(z\mid \bx_T,\by_T,\bx_C,\by_C)}[\log q_\phi(z\mid \bx_T,\by_T,\bx_C,\by_C) \nonumber\\
    & - \log p(z) - \log p(\by_T\mid \bx_T,\bx_C,\by_C,z) \nonumber\\
    & + \log p(\by_T\mid\bx_T,\bx_C,\by_C)] \geq 0 \label{eq:KL_expand_append}
\end{align}
Rearranging, we have
\begin{align}
    & \log p(\by_T\mid\bx_C,\by_C,\bx_T) \nonumber\\
    \geq & \EE_{z\sim q_\phi(z\mid \bx_T,\by_T,\bx_C,\by_C)}[\log p(\by_T\mid \bx_T,\bx_C,\by_C,z) \nonumber\\
    &- \log q_\phi(z\mid \bx_T,\by_T,\bx_C,\by_C) + \log p(z)] \nonumber\\
    = & \EE_{z\sim q_\phi(z\mid \bx_T,\by_T,\bx_C,\by_C)}[\log p(\by_T\mid \bx_T,\bx_C,\by_C,z)] \nonumber\\
    & - D_\mathsf{KL}(q_\phi(z\mid \bx_T,\by_T,\bx_C,\by_C) \mid\mid p(z)) \triangleq \mathsf{ELBO} \label{eq:ELBO_append}
\end{align}
Plug \eqref{eq:KL_expand_append} and \eqref{eq:ELBO_append} into \eqref{eq:KL_append}, we have
\begin{align}
    & \minimizewrt{\phi} D_{\mathsf{KL}}(q_\phi(z\mid \bx_T,\by_T,\bx_C,\by_C) \mid\mid p(z\mid \bx_T,\by_T,\bx_C,\by_C)) \nonumber \\
    \equiv & \minimizewrt{\phi} -\mathsf{ELBO} + \log p(\by_T\mid\bx_T,\bx_C,\by_C) \nonumber \\
    \equiv & \maximizewrt{\phi} \mathsf{ELBO}
\end{align}
since $\log p(\by_T\mid\bx_T,\bx_C,\by_C)$ is intractable and does not depend on $\phi$. Parameterizing the likelihood $p(\by_T\mid \bx_T,\bx_C,\by_C,z)$ in \eqref{eq:ELBO_append}, we arrive at \eqref{eq:elbo_1} and \eqref{eq:ELBO_def}. $\blacksquare$

% \resizebox{0.49\textwidth}{!}{
% \begin{tabular}{p{2cm}p{10cm}p{5cm}}
%     \centering
    
% \end{tabular}
% }

% \textbf{Final Output MLP $f_y$}: $[156, 128, 64, 2]$

% \textbf{Context att}: Key/query head: $[128, 32]$

% \textbf{Finger Feat}: This is separated into two stages. First, finger velocities $x^{(t)}$ are fed into an MLP of sizes $[6, 32, 32]$ to produce an features $x^{'(t)}$. These are then passed along with $h^{(t-1)}$ into a second stage MLP of sizes $[320, 128, 64, 32]$.

% \textbf{Latent Z MLP $q_{\phi}(z|\cdot)$}
% Each ($h^{(t)}, y^{(t)})$ pair is fed into an MLP of sizes $[]$ to produce features, which are averaged together to produce $s_C$. These are then fed into another MLP of sizes $[]$ to produce features. Such features are finally fed into two MLPs of the same sizes $[128, 32]$ to predict $\mu_z$ and $\sigma_z$ that parameterize the distribution over latent $z$.

% \textbf{LSTMCell}: Two adjacent LSTMCell with sizes $[230, 128]$ and $[128, 128]$

% \textbf{Final Output MLP $f_y$}: $[156, 128, 64, 2]$

\subsection{Train Test Data Collection}\label{append:train_test_data}
In section \ref{sec:collect_user_policies}, we mentioned that all $72$ collected data clips from $\mathsf{in-sample}$ users are separated into $D_\mathsf{train}$ and $D_\mathsf{test, in-sample}$ sets. Each clip can contain one or two active dimensions depending on whether translation or rotation or both are active. Out of these $72$ clips, $24$ of the clips will only contain one active dimension. These are split evenly among $D_\mathsf{train}$ and $D_\mathsf{test, in-sample}$. The remaining $48$ clips will contain two active dimensions each, and these are also evenly split among the two datasets. Among the $24$ of these assigned to $D_\mathsf{test, in-sample}$, half are comprised of active dimensions that are also captured in $D_\mathsf{train}$. So overall, $D_\mathsf{train}$ and $D_\mathsf{test, in-sample}$ will overall be given 36 clips each from each $\mathsf{in-sample}$ user.

% \ruic{How clips are constructed, how $D_\mathsf{train}$ and $D_\mathsf{test}$ are designed. Describe in corresponding to previous definitions.}

\subsection{Model Architecture}\label{append:model_arch}

% Additional topics to discuss:
% - Multi-head Context Attention, specifically one head for each dimension of motion. At a given timestep, only one or two dominant motions may be active, while for other timesteps, other dimensions may be active. Thus, different dimensions should place attention on different timesteps that are most relevant to their specific dimension.

% \paragraph*{\textbf{Long Short-Term Finger Attention Cell (FAC)}}

The configurations of different MLP modules introduced in section \ref{sec:arch} are given in table \ref{table:model_arch_config}. Each configuration starts with input size and ending with output size. A ReLU activation is applied after input layer and every hidden layer. Introduced in section \ref{sec:cchp_encoder}, the hidden state sizes $H$ and $H'$ are $128$ and $32$ respectively.

\begin{table}[h]\centering
    \begin{tabular}{lll}
        \toprule
        Module & Description & Configuration  \\
        \midrule
        $f_\mathsf{finger}$ &
        \begin{tabular}{@{}l@{}} Finger feature (\ref{sec:cchp_encoder}) \end{tabular}
         & [6, 32, 32]\\
        $f_\mathsf{hand}$ &
        \begin{tabular}{@{}l@{}} Hand feature (\ref{sec:cchp_encoder}) \end{tabular}
         & [320, 128, 64, 32]\\
        \midrule
        $f_a$ &
        \begin{tabular}{@{}l@{}} $\mathbf{h}$, $\mathbf{y}$ aggregation feature \eqref{eq:sC_fa} \end{tabular}
        & [134, 128, 128, 128, 128]\\
        \midrule
        $f_{\phi}$ & Latent posterior predictor \eqref{eq:f_phi} &
        \begin{tabular}{@{}l@{}} [128, 128, 128, 128, 128] \\ and two heads [128, 32] \end{tabular} \\
        \midrule
        $f_{\mathsf{kq}}$ &
        \begin{tabular}{@{}l@{}} Key query feature in \\ context attention \eqref{eq:context_att} \end{tabular}
        & [128, 32] \\
        \midrule
        $f_y$ & Operation prediction \eqref{eq:f_y} &   \begin{tabular}{@{}l@{}}[156, 128, 64, 2], one for \\ each motion dimension  \end{tabular} \\
        \bottomrule
    \end{tabular}
    \caption{MLP configurations for CCHP neural network implementation.}
    \label{table:model_arch_config}
\end{table}

\subsection{Model Training}\label{append:model_training}
We train each model using the Adam optimizer with learning rate $lr = 5e-4$. During training, command finger velocities $\bx$ are perturbed with Gaussian noises $\epsilon \sim \cN(0, 1e-6)$. We train our models for a total of $738$ epochs with batch size $32$, which took around $5$ hours using an Nvidia GeForce RTX 2080Ti GPU with an Intel i9-9940X CPU.

% \ruic{hyperparams}

% \ruic{computer spec, cpu, gpu, etc.}

% \subsection{Real-Time Cobot Handling (continued)}\label{append:demo}

% \ruic{More screenshots of hardware demo}

% \section*{ACKNOWLEDGMENT}

% xxx

%%%%%%%%%%%%%%%%%%%%%%%%%%%%%%%%%%%%%%%%%%%%%%%%%%%%%%%%%%%%%%%%%%%%%%%%%%%%%%%%

% \clearpage
\bibliographystyle{IEEEtran}
% \bibliography{IEEEabrv,IEEE}
\bibliography{ref}

\end{document}